\documentclass[10pt,journal,compsoc]{IEEEtran}
\usepackage[table]{xcolor}
\usepackage{arydshln}
\usepackage{bm}
\usepackage{amsfonts}
\usepackage{amsmath}
\usepackage{subcaption}
\usepackage{makecell}
\definecolor{themebrown}{RGB}{161, 120, 73}
\definecolor{themeblue}{RGB}{57, 162, 219}
\definecolor{themegreen}{RGB}{87, 204, 153}
\definecolor{forestgreen}{RGB}{47, 159, 87}
\definecolor{link}{RGB}{75, 166, 154}

\definecolor{good}{RGB}{47, 159, 87}
\definecolor{bad}{RGB}{0, 0, 0}

\definecolor{convnet}{RGB}{47, 159, 87}
\definecolor{vit}{RGB}{144, 113, 74}

\newcommand{\vb}{{\color{vit}{$\bm{\circ}$\,}}}
\newcommand{\cb}{{\color{convnet}{$\bullet$\,}}}
\newcommand{\nb}{{\color{white}{$\bullet$\,}}}

\usepackage[colorlinks=true, allcolors=forestgreen]{hyperref}
\newcommand{\cmark}{\color{good}\ding{51}}%
\newcommand{\xmark}{\color{bad}\ding{55}}%

\usepackage{multirow}
\usepackage{pifont}
\usepackage{graphicx}
\usepackage{xspace}
\usepackage{ragged2e}
\usepackage{makecell}
\usepackage{booktabs}

\newcommand{\tablestyle}[2]{\setlength{\tabcolsep}{#1}\renewcommand{\arraystretch}{#2}\centering\small}

\newcommand\gray[1]{\color{gray}#1}
\newcommand\graycell[0]{\cellcolor{midgrey}}
\def\x{$\times$}
\definecolor{midgrey}{RGB}{235,235,235}

\newlength\savewidth\newcommand\shline{\noalign{\global\savewidth\arrayrulewidth
  \global\arrayrulewidth 1pt}\hline\noalign{\global\arrayrulewidth\savewidth}}

\ifCLASSOPTIONcompsoc
  \usepackage[nocompress]{cite}
\else
  \usepackage{cite}
\fi
\ifCLASSINFOpdf
\else
\fi
\begin{document}
\title{Temporally-Adaptive Models \\ for Efficient Video Understanding}

\author{Ziyuan Huang, Shiwei Zhang*, Liang Pan, Zhiwu Qing, Yingya Zhang, Ziwei Liu, Marcelo H. Ang Jr*
\IEEEcompsocitemizethanks{
\IEEEcompsocthanksitem * Correspondence to Shiwei Zhang (zhangjin.zsw@alibaba-inc.com) and Marcelo H. Ang Jr (mpeangh@nus.edu.sg).
\IEEEcompsocthanksitem Ziyuan Huang and Marcelo H. Ang Jr are with Advanced Robotics Centre, National University of Singapore.
\IEEEcompsocthanksitem Shiwei Zhang and Yingya Zhang is with DAMO Academy, Alibaba Group.
\IEEEcompsocthanksitem Liang Pan and Ziwei Liu are with S-Lab, National Technological University.
\IEEEcompsocthanksitem Zhiwu Qing is with Key Laboratory of Image Processing and Intelligent Control, School of Artificial Intelligence and Automation, Huazhong University of Science. }
}

%
%

\markboth{Journal of \LaTeX\ Class Files,~Vol.~14, No.~8, August~2015}%
{Shell \MakeLowercase{\textit{et al.}}: Bare Demo of IEEEtran.cls for Computer Society Journals}
%



\IEEEtitleabstractindextext{%
\begin{abstract}
\justifying
Spatial convolutions\footnote{In this work, we use spatial convolutions and 2D convolutions interchangeably.} are extensively used in numerous deep video models.
It fundamentally assumes spatio-temporal invariance, \textit{i.e.}, using shared weights for every location in different frames.
This work presents Temporally-Adaptive Convolutions (TAdaConv) for video understanding, which shows that adaptive weight calibration along the temporal dimension is an efficient way to facilitate modeling complex temporal dynamics in videos.
Specifically, TAdaConv empowers spatial convolutions with temporal modeling abilities by calibrating the convolution weights for each frame according to its local and global temporal context.
Compared to existing operations for temporal modeling, TAdaConv is more efficient as it operates over the convolution kernels instead of the features, whose dimension is an order of magnitude smaller than the spatial resolutions.
Further, kernel calibration brings an increased model capacity.
Based on this readily plug-in operation TAdaConv as well as its extension, \textit{i.e.,} TAdaConvV2, we construct TAdaBlocks to empower ConvNeXt and Vision Transformer to have strong temporal modeling capabilities. 
Empirical results show TAdaConvNeXtV2 and TAdaFormer perform competitively against state-of-the-art convolutional and Transformer-based models in various video understanding benchmarks. Our codes and models are released at: \url{https://github.com/alibaba-mmai-research/TAdaConv}.
\end{abstract}

}

\maketitle

\IEEEdisplaynontitleabstractindextext

%
\IEEEpeerreviewmaketitle

\section{Introduction}\label{sec:introduction}
Convolutions are an indispensable operation in modern deep vision models~\cite{resnet,inception,alexnet,dai2021coatnet}, whose different variants have driven the state-of-the-art performances of convolutional neural networks (CNNs) in many visual tasks~\cite{resnext,deformable,nes,zhang2022resnest,tian2020condinst} and application scenarios~\cite{mobilenets,condconv}. 
In the video paradigm, compared to the 3D convolutions~\cite{c3d}, the combination of 2D spatial convolutions and 1D temporal convolutions is more widely preferred owing to its efficiency~\cite{r21d,p3d}.
Nevertheless, 1D temporal convolutions introduce non-negligible computation overhead on top of the spatial convolutions. Therefore, we seek to directly equip spatial convolutions with temporal modeling abilities.

One essential property of convolutions is the translation invariance~\cite{ruderman1994statistics,simoncelli2001natural}, resulting from its local connectivity and shared weights. 
However, recent works in dynamic filtering have shown that strictly shard weights for all pixels may be sub-optimal for modeling various spatial contents~\cite{ddf,wu2018dynamic}.

Given the diverse nature of the temporal dynamics in videos, we hypothesize that \textit{temporal modeling could benefit from relaxed invariance along the temporal dimension}. This means that convolution weights for different time steps are no longer strictly shared.
Existing dynamic filter networks could achieve this but with two drawbacks. \textbf{(i)} it is difficult for most of them~\cite{ddf,condconv} to leverage pre-trained weights, which is critical in video applications since training video models from scratch is highly resource demanding~\cite{slowfast,x3d} and prone to over-fitting on small datasets.
\textbf{(ii)} for most dynamic filters, the weights are generated with respect to its spatial context~\cite{ddf,dynamicfilter} or the global descriptor~\cite{dycnn,condconv}, which is incapable of capturing the fine-grained temporal variations between frames.

\begin{figure*}[t]
\centering
\includegraphics[width=0.85\linewidth]{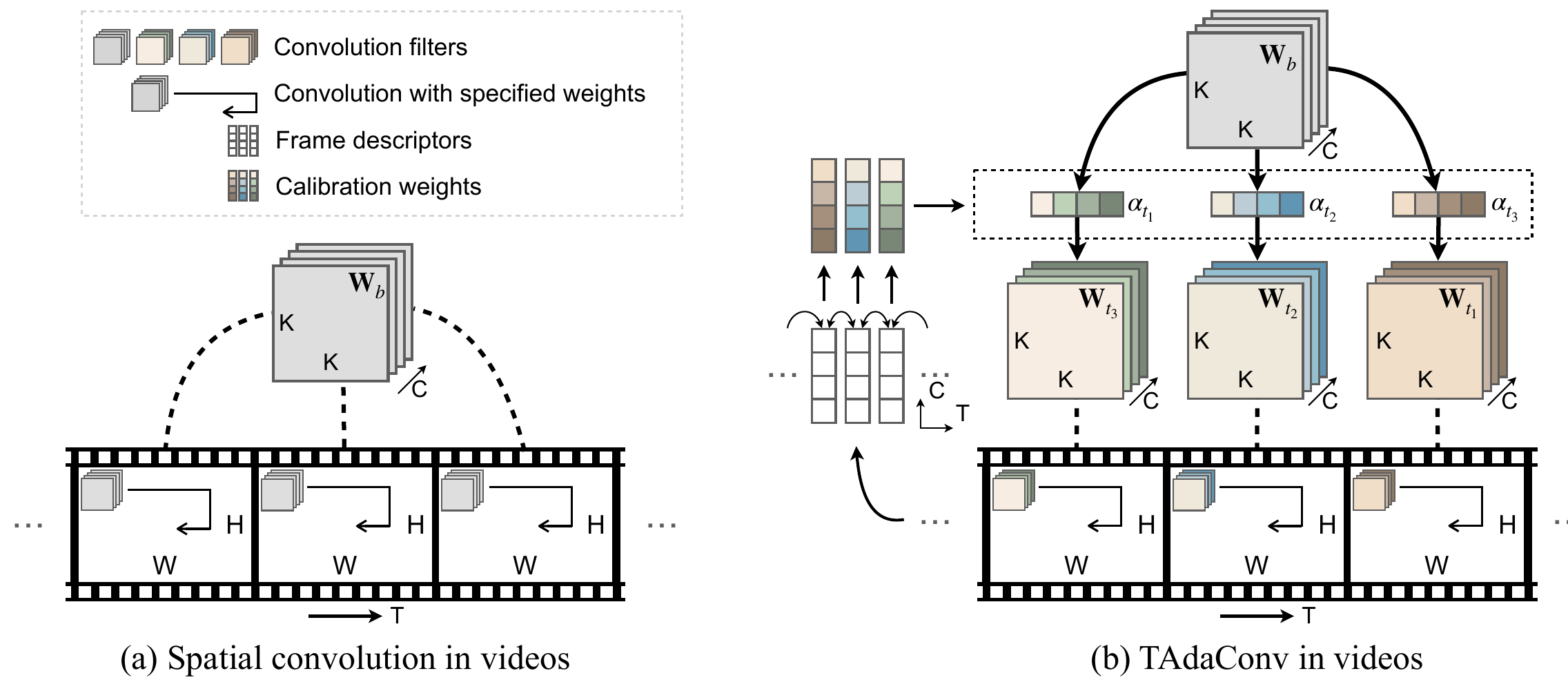}
\vspace{-3mm}
\caption{\textbf{Comparisons between TAdaConv and the spatial convolutions in video models}. 
(a) Standard spatial convolutions in videos share the kernel weights between different frames. 
(b) Our TAdaConv adaptively calibrates the kernel weights for each frame by its temporal context.
}
\vspace{-5mm}
\label{fig:spatialconvcomp}
\end{figure*}

Motivated by this, 
we present Temporally-Adaptive Convolution (TAdaConv) for video understanding, where the convolution weights are no longer fixed across different frames. 
Specifically, the convolution kernel for the $t$-th frame $\mathbf{W}_t$ is factorized to the multiplication of the base weight and a calibration weight: $\mathbf{W}_t = \bm{\alpha}_t \cdot \mathbf{W}_b$, where the base weight $\mathbf{W}_b$ is learnable and the calibration weight $\bm{\alpha}_t$ is adaptively generated from the input data in the base weight $\mathbf{W}_b$. 
For each frame, we generate the calibration weight based on the frame descriptors of its adjacent time steps as well as the global descriptor, which effectively encodes both local and global temporal dynamics in videos. The difference between TAdaConv and standard convolutions is visualized in Fig.~\ref{fig:spatialconvcomp}.

The main advantages of this factorization are three-fold: \textbf{(i)} TAdaConv can be easily plugged into any existing models to enhance temporal modeling, and their pre-trained weights can still be exploited; \textbf{(ii)} the temporal modeling ability can be highly improved with the help of the temporally-adaptive weight; \textbf{(iii)} in comparison with temporal convolutions that often operate on the learned 2D feature maps, TAdaConv is more efficient by directly operating on the convolution kernels.

TAdaConv is proposed as a drop-in replacement for the convolutions in existing models. 
A preliminary version of this work~\cite{huangtada} is published in ICLR 2022, where TAdaConv has demonstrated a strong capability of temporal modeling, introducing notable performance gains to both image-based models as well as existing video models. 
In this work, we follow the conceptual idea of TAdaConv and present improvements to the preliminary version on both structural designs as well as model and data scaling. In terms of structural designs, we optimize TAdaConv in the following aspects:
\textbf{(i)} At the operation level, the calibration factor generation process of TAdaConv is optimized, where multi-head self-attention~\cite{vaswani2017attention} is introduced for modeling the global information of the videos. 
\textbf{(ii)} At the block level, we construct stronger TAdaBlocks by introducing efficient temporal feature aggregation, which we use to construct our convolutional model TAdaConvNeXtV2 and transformer TAdaFormer.
Our empirical results show a notable improvement brought by our modifications on both scene- and motion-centric benchmarks. 
Based on the TAdaConvNeXtV2 and TAdaFormer, we further scale up both the model and data scale, which lead to a competitive performance to existing state-of-the-art approaches.

\begin{figure*}[t]
\centering
\includegraphics[width=0.9\textwidth]{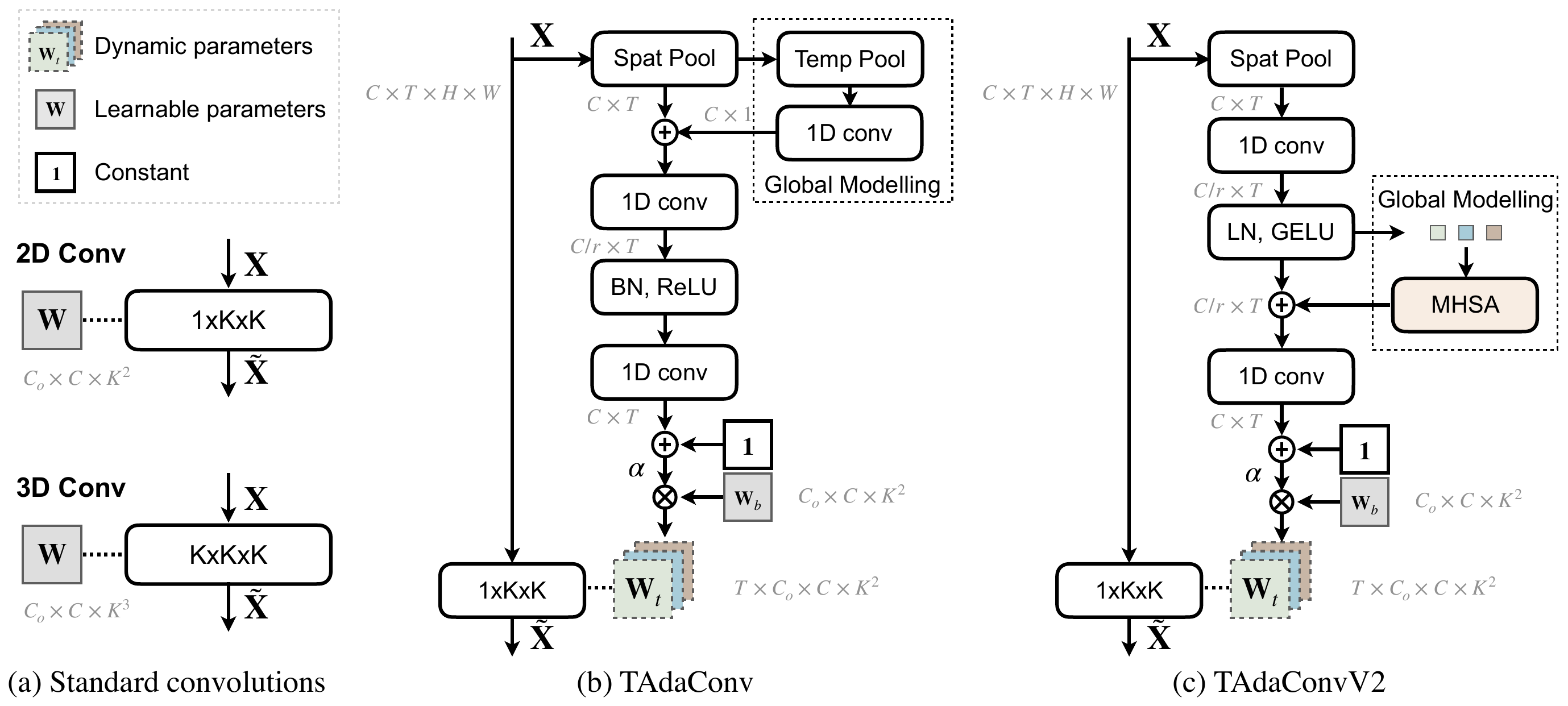}
\caption{
\textbf{Instantiations of TAdaConv and TAdaConvV2. }
(a) Standard convolutions used in video models. 
(b) Our TAdaConv using non-linear weight calibrations with global temporal context.
(c) TAdaConvV2 with global temporal context provided by multi-head self-attention (MHSA).
}
\vspace{-4mm}
\label{fig:tadaconvinstantiation}
\end{figure*}

\section{Related Work}

\textbf{Convolutional models for video understanding. }
Early convolutional models obtain spatio-temporal representations by 3D convolutions~\cite{i3d,c3d,x3d,csn} or two-stream networks~\cite{twostream}.
For efficiency, recent ones build upon 2D networks and design additional operations for temporal modeling~\cite{tsm,tdn,stm,r21d,tam,corrnet,wang2023e3d,smallbignet,zhou2023gcm}, where the weights of the 2D convolutions are shared between different timestamps. 
Our preliminary version~\cite{huangtada} find removing this constraint leads to stronger temporal modeling ability. 
In this work, we modernize the convolutional model according to ConvNeXt~\cite{convnext} and construct a stronger convolutional model for video understanding.

\textbf{Vision Transformers for video understanding.} 
With the great success of Transformers in natural language processing~\cite{vaswani2017attention,kenton2019bert,radford2018gpt}, Vision Transformers (ViT)~\cite{dosovitskiy2020vit} are showing strong performances in various vision tasks~\cite{mvit,meinhardt2022trackformer,cao2022tctrack,bao2021beit,zhou2022pttr,ramesh2022dalle2,carion2020detr} including video understanding~\cite{nonlocal,videoswin,arnab2021vivit,timesformer,liu2022tadtr,chen2023videollm,patrick2021motionformer}.
The capability of ViTs is further enhanced when it is pre-trained on a large corpus of image~\cite{chen2021mocov3,he2022mae}, video~\cite{tong2022videomae,feichtenhofer2022maest,qing2023hico} or multi-modal data~\cite{radford2021clip,yu2022coca}, or when the size of the model is increased~\cite{zhai2022scaling,dehghani2023scaling22b}, or both~\cite{wang2022beitv3,wang2023videomaev2}. 
Since directly pre-training with video data is both resource- and time-consuming, an alternative is to exploit the models pre-trained on large-scale image data and empower the model thorough additional structures for temporal modeling, such as temporal ~\cite{timesformer} or 3D windowed self-attention~\cite{videoswin}, spatio-temporal adapters~\cite{pan2022stadapter}, \textit{etc}.
In our work, we exploit the vanilla Vision Transformer pre-trained on a large corpus of image-text data~\cite{radford2021clip} and equip it with strong temporal modeling ability with our TAdaBlock.

\textbf{Dynamic networks. }
Dynamic networks refer to networks with content-adaptive weights or modules, such as dynamic filters/convolutions~\cite{dynamicfilter,condconv,dcd,ddf},
dynamic activations~\cite{micronet,dynamicrelu}, and dynamic routing~\cite{skipnet,dynamicrouting}, \textit{etc}.
They have demonstrated exceeding network capacity and performance compared to static ones in various tasks~\cite{ye2022dck,jiang2020convbert,xu2020udvd,wu2019dynamicconv} as well as in video understanding~\cite{tam,adafuse,wu2019adaframe,meng2020arnet}.
Some recent spatially-adaptive convolutions~\cite{lrlc,drconv} show relaxing spatial invariance could help modeling diverse visual contents, and our preliminary version~\cite{huangtada} shows video understanding can benefit from relaxing the temporal invariance.
This work further exploits the idea and enhance the temporal modeling capability of TAdaConv by introducing multi-head self-attention for global temporal modeling.

\section{Temporally-adaptive Convolutions}

In this work, we seek to empower the spatial convolutions with temporal modeling abilities. 
Inspired by the calibration process of temporal convolutions (Sec.~\ref{Sec:TempConv}), TAdaConv dynamically calibrates the convolution weights for each frame (Sec.~\ref{Sec:TempVariance}) according to its temporal context (Sec.~\ref{Sec:Calibration_weight_generation}). 

\subsection{Revisiting temporal convolutions}
\label{Sec:TempConv}

We first revisit the temporal convolution to show the underlying process and its relation to dynamic filters. 
We consider depth-wise temporal convolution for simplicity, which is more widely used due to its efficiency~\cite{tam,stm}.
Formally, for a 3\x1\x1 temporal convolution filter parameterized by $\bm{\beta}=[\bm{\beta}_1, \bm{\beta}_2, \bm{\beta}_3]$ and placed (ignoring normalizations) after the 2D convolution parameterized by $\mathbf{W}$, the output feature $\mathbf{\tilde{x}}_t$ of the $t$-th frame can be obtained by:
\begin{equation}
    \mathbf{\tilde{x}}_t = \bm{\beta}_1 \cdot \delta(\mathbf{W} * \mathbf{x}_{t-1}) + \bm{\beta}_2 \cdot \delta(\mathbf{W} * \mathbf{x}_{t}) + \bm{\beta}_3 \cdot \delta(\mathbf{W} * \mathbf{x}_{t+1})\ ,
    \label{eq:tempspatconv}
\end{equation}
\noindent where the $\cdot$ indicates the element-wise multiplication, $*$ denotes the convolution over the spatial dimension and $\delta$ denotes ReLU activation~\cite{relu}. It can be rewritten as follows:
\begin{equation}
    \mathbf{\tilde{x}}_t = \mathbf{W}_{t-1} * \mathbf{x}_{t-1} + \mathbf{W}_{t} * \mathbf{x}_{t} + \mathbf{W}_{t+1} * \mathbf{x}_{t+1}\ ,
    \label{eq:simplerform_tempspatconv}
\end{equation}
\noindent where $\mathbf{W}_{t-1}^{i,j}=\mathbf{M}_{t-1}^{i,j}\cdot\bm{\beta}_1\cdot\mathbf{W}, \mathbf{W}_t^{i,j}=\mathbf{M}_{t}^{i,j}\cdot\bm{\beta}_2\cdot\mathbf{W}$ and $\mathbf{W}_{t+1}^{i,j}=\mathbf{M}_{t+1}^{i,j}\cdot\bm{\beta}_3\cdot\mathbf{W}$ are spatio-temporal location adaptive convolution weights. $\mathbf{M}_t\in\mathbb{R}^{C\times H\times W}$ is a dynamic tensor, with its value dependent on the result of the spatial convolutions (see \textit{Appendix} for details). Hence, the temporal convolutions in the (2+1)D convolution essentially perform \textbf{(i)} weight calibration on the spatial convolutions and \textbf{(ii)} feature aggregation between adjacent frames. 
However, if the temporal modeling is achieved by coupling temporal convolutions to spatial convolutions, a non-negligible computation overhead is still introduced (see Table~\ref{tab:temporalmodelingcomparison}). 

\subsection{Formulation of TAdaConv and TAdaConvV2}
\label{Sec:TempVariance}
For efficiency, we set out to directly empower the spatial convolutions with temporal modeling abilities. 
Inspired by the recent finding that the relaxation of spatial invariance strengthens spatial modeling~\cite{ddf,lrlc}, we hypothesize that temporally adaptive weights can also help temporal modeling. 
Therefore, the convolution weights in a TAdaConv layer are varied on a frame-by-frame basis.
Since we observe that previous dynamic filters can hardly utilize the pretrained weights, we take inspiration from our observation in the temporal convolutions and factorize the weights for the $t$-th frame $\mathbf{W}_t$ into the multiplication of a base weight $\mathbf{W}_b$ shared for all frames, and a calibration weight $\bm{\alpha}_t$ that are different for each time step:
\begin{equation}
    \mathbf{\tilde{x}}_t = \mathbf{W}_t * \mathbf{x}_t = (\bm{\alpha}_t \cdot \mathbf{W}_b) * \mathbf{x}_t\ .
\end{equation}
\subsection{Calibration weight generation. }
\label{Sec:Calibration_weight_generation}
To allow for the TAdaConv to model temporal dynamics, it is crucial that the calibration weight $\bm{\alpha}_t$ for the $t$-th frame takes into account not only the current frame, but more importantly, its temporal context, \textit{i.e.,} $\bm{\alpha}_t = \mathcal{G}(...,\mathbf{x}_{t-1},\mathbf{x}_{t},\mathbf{x}_{t+1},...)$.
Otherwise, TAdaConv would degenerate to a set of unrelated spatial convolutions with different weights applied on different frames.
In practice, the calibration generation function can have various structural designs. In Fig.~\ref{fig:tadaconvinstantiation}(b) and (c), we show two instantiations of the calibration generation function, which respectively correspond to TAdaConv and TAdaConvV2. 

\noindent\textbf{TAdaConv.} In our design, we aim for efficiency and the ability to capture inter-frame temporal dynamics. 
For efficiency, we operate on the frame description vectors $\mathbf{v}\in\mathbb{R}^{T\times C}$ obtained by the global average pooling over the spatial dimension $\text{GAP}_s$ for each frame, \textit{i.e.,} $\mathbf{v}_t=\text{GAP}_s(\mathbf{x}_t)$.
For temporal modeling, we apply two-layer 1D convolutions $\mathcal{F}$ with a dimension reduction ratio of $r$ on the local temporal context $\mathbf{v}_t^{adj} = \{\mathbf{v}_{t-1}, \mathbf{v}_t, \mathbf{v}_{t+1}\}$:
\begin{equation}
\begin{aligned}
    \mathbf{v}_t^{\prime adj} &= \text{ReLU}(\text{BN}(f^{C\rightarrow C/r}(\mathbf{v}_t^{adj})))\\
    \mathcal{F}(\mathbf{v}_t) &=  f^{C/r\rightarrow C}(\mathbf{v}_t^{\prime adj})
\end{aligned}\ ,
\end{equation}
\noindent where we use ReLU~\cite{relu} and batch normalizations~\cite{bn} for activation and normalization. $f$ denotes 1-D convolutions. 

In order for a larger inter-frame field of view in complement to the local 1D convolution, we further incorporate global temporal information into the calibration weight generation process. For TAdaConv, we add a global descriptor 
to the weight generation process $\mathcal{F}$ through a linear mapping function $\text{FC}$:
\begin{equation}
\begin{aligned}
    \mathbf{v}_t^{\prime adj} &= \mathbf{v}_t^{adj} + \text{FC}(\text{GAP}_t(\mathbf{v}_t)) \\
    \mathbf{v}_t^{\prime\prime adj} &= \text{ReLU}(\text{BN}(f^{C\rightarrow C/r}(\mathbf{v}_t^{\prime adj}))\\
    \mathcal{F}(\mathbf{v}_t) &=  f^{C/r\rightarrow C}(\mathbf{v}_t^{\prime\prime adj})
\end{aligned}\ ,
\end{equation}
\noindent where $\text{GAP}_t(\mathbf{v}_t)$ denotes global average pooling over the temporal dimension on the frame descriptors $\mathbf{v}_t$. This is equivalent to global average pooling over all spatiotemporal dimensions on the original input $\mathbf{x}$. Hence, $\text{GAP}_t(\mathbf{v}_t)$ contains the global temporal context in the input videos.

\textbf{TAdaConvV2. }The instantiation of TAdaConvV2 is generally similar to TAdaConv, with two improvements. \textbf{(i)} We alter the combination of ReLU and batch normalizations to GELU and layer normalizations to conform to the structures in ConvNeXt models. \textbf{(ii)} For global temporal context modeling, we take advantage of the powerful global modeling capability of self-attention~\cite{vaswani2017attention}. Specifically, the calibration weight generation function can be expressed as follows:
\begin{equation}
\begin{aligned}
    \mathbf{v}_t^{\prime adj} &= \text{GELU}(\text{LN}(f^{C\rightarrow C/r}(\mathbf{v}_t^{adj})))\\
    \mathbf{v}_t^{\prime\prime adj} &= \text{MHSA}(\mathbf{v}_t^{\prime adj}) + \mathbf{v}_t^{\prime adj}\\
    \mathcal{F}(\mathbf{v}_t) &= f^{C/r\rightarrow C}(\mathbf{v}_t^{\prime\prime adj})
\end{aligned}\ ,
\end{equation}
\noindent where MHSA denotes the multi-head self-attention~\cite{vaswani2017attention}. Since the 1D convolution before MHSA essentially provides a dynamic positional embedding for the frame descriptors $\mathbf{v}$, we do not add additional positional embeddings before the MHSA operation.

\begin{table}[b]
\caption{
Comparison of (2+1)D convolution and TAdaConv in FLOPs and number of parameters. 
Example setting for the operation: $C_o=C_i=64$, $K=3$, $T=8$, $H=W=56$ and $r=4$. 
Example setting for the network: ResNet-50 with input resolution $8\times 224^2$.
Colored numbers denote the extra FLOPs/parameters introduced to 2D convolutions or ResNet-50.
Refer to \textit{Appendix} for model structures.
}
\centering
\tablestyle{3pt}{1.0}
\begin{tabular}{lll}
\shline
 ~& \textbf{(2+1)D Conv} & \textbf{TAdaConv} \\
\hline
\multirow{4}{*}{FLOPs} & ~ & \footnotesize $C_o\times C_i\times K^2\times THW$  \\
~ & \footnotesize $C_o\times C_i\times K^2 \times THW$ & \footnotesize $+ C_i\times (THW+T)$ \\
~ & \footnotesize $+C_o\times C_i\times K \times THW$ & \footnotesize $+ C_i\times C_i/r\times (2\times K \times T+1)$ \\
 ~ & ~ & \footnotesize $+C_o\times C_i\times K^2 \times T$ \\
E.G. Op & 1.2331 {\footnotesize\color{bad}\textbf{(+0.308, $\uparrow$33$\%$)}} & 0.9268 {\footnotesize\color{forestgreen}\textbf{(+0.002, $\uparrow$0.2$\%$)}}\\
E.G. Net & 37.94 {\footnotesize\color{bad}\textbf{(+4.94, $\uparrow$15$\%$)}} & 33.02 {\footnotesize\color{forestgreen}\textbf{(+0.02, $\uparrow$0.06$\%$)}} \\
\hline
\multirow{2}{*}{Params.} & \footnotesize $C_o\times C_i\times K^2$  & \footnotesize $C_o\times C_i\times K^2$\\
~ & \footnotesize $+C_o\times C_i\times K$ & \footnotesize $+2\times C_i\times C_i/r\times K$ \\

E.G. Op.& 49,152 {\footnotesize\color{bad}\textbf{(+12,288, $\uparrow$33$\%$)}} & 43,008 {\footnotesize\color{forestgreen}\textbf{(+6,144, $\uparrow$17$\%$)}}\\
E.G. Net & 28.1M {\footnotesize\color{bad}\textbf{(+3.8M, $\uparrow$15.6$\%$)}}& 27.5M {\footnotesize\color{forestgreen}\textbf{(+3.2M, $\uparrow$13.1$\%$)}}\\
\shline
\end{tabular}
\label{tab:temporalmodelingcomparison}
\end{table}

\textbf{Initialization. }The TAdaConv is designed to be readily inserted into existing models by simply replacing the 2D convolutions. 
For effective use of the pre-trained weights, TAdaConv is initialized to behave exactly the same as the standard convolution.
This is achieved by zero-initializing the weight of the last convolution in $\mathcal{F}$ and adding a constant vector $\mathbf{1}$ to the formulation:
\begin{equation}
    \bm{\alpha}_t = \mathcal{G}(\mathbf{x}) = \mathbf{1} + \mathcal{F}(\text{GAP}_s(\mathbf{x}_t^{adj}))\ .
\end{equation}
\noindent In this way, at initial state, $\mathbf{W}_t=\mathbf{1}\cdot\mathbf{W}_b=\mathbf{W}_b$, where we load $\mathbf{W}_b$ with the pre-trained weights.

\textbf{Calibration dimension. }
The base weight $\mathbf{W}_b\in\mathbb{R}^{C_{\text{out}}\times C_{\text{in}}\times K^2}$ can be calibrated in different dimensions.
For standard convolutions, we instantiate the calibration on the $C_{\text{in}}$ dimension ($\bm{\alpha}_t\in\mathbb{R}^{1\times C_{\text{in}}\times1}$), as the weight generation based on the input features yields a more precise estimation for the relation of the input channels than the output channels or spatial structures (empirical analysis in Table~\ref{tab:calibrationdim}). For depthwise convolutions, since the convolution kernel does not have a $C_{\text{in}}$ dimension, the calibration is directly applied on the $C_{out}$ dimension of the convolution kernel.

\textbf{Comparison with temporal convolutions.} Table~\ref{tab:temporalmodelingcomparison} compares the TAdaConv with R(2+1)D in parameters and FLOPs, which shows most of our additional computation overhead on top of the spatial convolution is an order of magnitude less than the temporal convolution. 

\begin{table}[b]
\tablestyle{5pt}{1.0}
\caption{Comparison with existing dynamic filters in terms of temporal modeling capability, location adaptiveness and the ability to exploit pre-trained weights in existing models.}
\label{tab:compdyconv}
\centering
\begin{tabular}{lccc}
\shline
~ &  \textbf{Temporal} &  \textbf{Location} & \textbf{Pretrained}\\
\textbf{Operations} &  \textbf{modeling} &  \textbf{adaptive} & \textbf{weights}\\
\hline
 CondConv~\cite{condconv} & \xmark & \xmark & \xmark \\
 DynamicFilter~\cite{dynamicfilter} & \xmark & \xmark & \xmark \\
 DDF~\cite{ddf} & \xmark & \cmark & \xmark \\
 TAM~\cite{tam} & \cmark & \xmark & \xmark \\
 \graycell TAdaConv(V2) & \graycell\cmark & \graycell\cmark & \graycell\cmark \\
\shline
\end{tabular}
\end{table}

\textbf{Comparison with existing dynamic filters.} Table~\ref{tab:compdyconv} compares TAdaConv with existing dynamic filters. 
The main difference between different dynamic filtering approaches lies in the way that the dynamic weights are generated. 
Mixture-of-experts-based dynamic filters~\cite{condconv} generate content-dependent weights to dynamically aggregate learnable convolution weights.
Other types of dynamic filters~\cite{dynamicfilter,ddf,tam} generate dynamic weights entirely based on the input content. 
Our TAdaConv is different from existing dynamic filters in the following three aspects:
\textbf{(i)} Compared to image-based dynamic filters~\cite{dynamicfilter,ddf,condconv}, TAdaConv achieves temporal modeling by generating weights based on the local and global context. 
\textbf{(ii)} Compared to TANet~\cite{tam} in the video paradigm, TAdaConv could model more complex temporal dynamics because of the temporally adaptive weights. 
\textbf{(iii)} Most existing dynamic filters are incapable of exploiting existing pre-trained weights, while TAdaConv could be initialized to generate dynamic weights that are identical to pre-trained ones. This reduces the training difficulty in video applications.
More detailed comparisons of dynamic filters are included in \textit{Appendix}.

\begin{figure*}[t]
\centering
\includegraphics[width=0.9\textwidth]{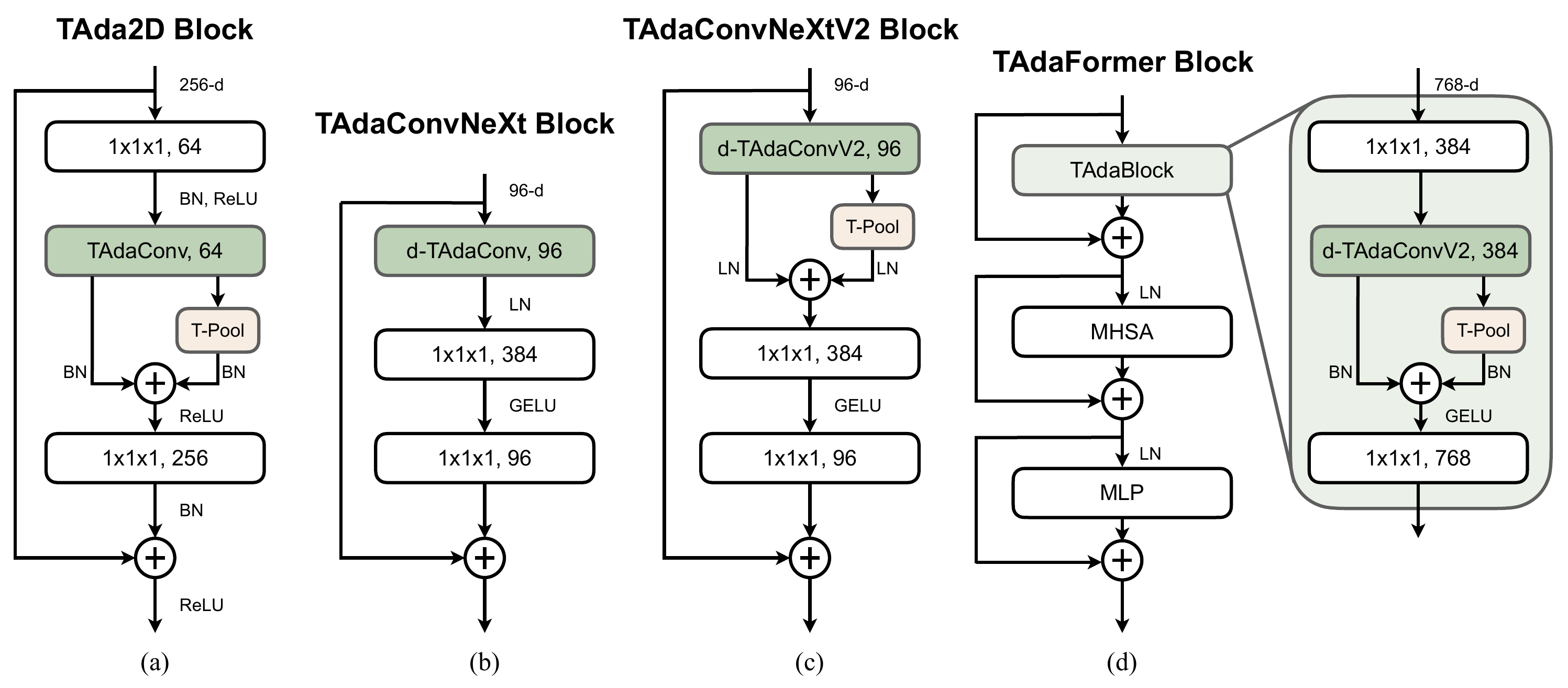}
\vspace{-4mm}
\caption{
\textbf{TAdaBlock designs for both convolutional and transformer-based models.} For convolutional models, we constructed TAda2D block in (a) and TAdaConvNeXt block in (b) in our preliminary version~\cite{huangtada}. Inspired by the efficient temporal aggregation in TAda2D, we introduce a similar strided temporal pooling (T-Pool in the figure) and a separate normalization layer to the TAdaConvNeXtV2 block in (c). For transformer-based models, we insert a TAdaBlock before each multi-head self-attention operation in the TAdaFormer block in (d), where the TAdaBlock is composed of two pointwise convolutions, a depth-wise TAdaConvV2, and a temporal feature aggregation module. 
}
\vspace{-2mm}
\label{fig:tadaconv-all-block}
\end{figure*}

\section{TAdaBlocks}
\label{Sec:TempAgg}
Based on TAdaConv and TAdaConvV2, we can construct a series of TAdaBlocks for various models, both convolutional and Transformer-based ones. In Fig.~\ref{fig:tadaconv-all-block}, we construct TAda2D block, TAdaConvNeXt(V2) block, and TAdaFormer block, respectively for ResNet~\cite{resnet}, ConvNeXt~\cite{convnext} and ViT~\cite{dosovitskiy2020vit}.

Apart from TAdaConv and TAdaConvV2, an important component of our TAdaBlocks is an efficient temporal feature aggregation scheme. This corresponds to the second essential step of temporal convolution. 
Formally, given the output of TAdaConv $\mathbf{\tilde{x}}$, the aggregated feature can be obtained as follows:
\begin{equation}
    \label{eq:aggregation}
    \mathbf{x}_{aggr} = \text{ReLU}( \text{Norm}_1(\mathbf{\tilde{x}}) + \text{Norm}_2(\text{T-Pool}_k(\mathbf{\tilde{x}})))\ ,
\end{equation}
\noindent where $\text{T-Pool}_k$ represents the strided temporal pooling operation with a kernel size of $k$. 
We use different normalization parameters for the features extracted by TAdaConv $\mathbf{\tilde{x}}$ and aggregated by strided average pooling $\text{TempAvgPool}_k(\mathbf{\tilde{x}})$, as their distributions are essentially different. 

During initialization, we load pre-trained weights (if any) to $\text{Norm}_1$, and initialize the parameters of $\text{Norm}_2$ to zero. 
Coupled with the initialization of TAdaConv, the initial state of the TAdaBlocks is exactly the same as the base model, while the calibration and the aggregation notably increase the model capacity with training (See \textit{Appendix}).
In experiments, we refer to this structure as the shortcut (Sc.) branch and the separate BN (SepBN.) branch.

In our preliminary version~\cite{huangtada}, we explored the TAda2D block and TAdaConvNeXt block. Inspired by the improvements brought by the temporal feature aggregation, we present an improved version of the TAdaConvNeXt block in Fig.~\ref{fig:tadaconv-all-block} (c), \textit{i.e.,} TAdaConvNeXtV2 block. 
To cater to the modernized convolutional block~\cite{convnext}, the structure of the aggregation scheme in TAdaConvNeXtV2 block is modified accordingly, where the activation function is removed and the normalization is switched to LayerNorm~\cite{ba2016layernorm}.

For Transformer-based models, we construct a TAdaFormer block, as in Fig.~\ref{fig:tadaconv-all-block} (d), where a ResNet-like convolutional block is inserted before each self-attention layer. Different from ResNet blocks, we use depth-wise TAdaConvV2 between two point-wise convolutions for efficiency. Inspired by the modernized convolutional block~\cite{convnext}, some of the normalization and activation layers are removed, as in Fig.~\ref{fig:tadaconv-all-block} (d). Temporal aggregation is similarly performed using the efficient feature aggregation scheme presented above. Empirically, we found batch normalizations work better in TAdaBlock for TAdaFormer. 

\section{Evaluations on video classification}

\textbf{Model.} 
We construct different variants for TAda2D, TAdaConvNeXtV2, and TAdaFormer, following the structure of the respective base models ResNet~\cite{resnet}, ConvNeXt~\cite{convnext}, and Vision Transformer~\cite{dosovitskiy2020vit}. Our model variants are obtained by replacing the residual blocks or the transformer blocks in the original model with our TAdaBlocks. Additionally, for TAdaConvNeXtV2 and TAdaFormer, we follow recent works~\cite{videoswin,arnab2021vivit} and use tubelet embedding stem. More details on the model structure is included in \textit{Appendix}.

\textbf{Datasets.}
For video classification, we use Kinetics-400~\cite{kinetics400} (\textit{K400}), Something-Something-V1 and V2~\cite{ssv2} (\textit{SSV1} and \textit{SSV2}), Epic-Kitchens-100~\cite{ek100} (\textit{EK100}),  and HACS~\cite{hacs}. 
Further, we employ UCF101~\cite{soomro2012ucf101} and HMDB51~\cite{kuehne2011hmdb} for multi-modal zero-shot evaluations.
\textit{K400} is a widely used action classification dataset with 400 categories covered by $\sim$300K videos.
\textit{SSV1} and \textit{SSV2} include 108K and 220K videos with challenging spatio-temporal interactions in 174 classes.
\textit{EK100} includes 90K segments labelled by 97 verb and 300 noun classes with actions defined by the combination of nouns and verbs. 
\textit{HACS} contains 504K videos with a taxonomy of 200 action classes. The latter two datasets are used for evaluation on action localization as well. 

In addition, we also construct a large-scale video classification dataset combining Kinetics-400~\cite{kinetics400}, Kinetics-600~\cite{carreira2018k600}, and Kinetics-700~\cite{carreira2019k700} for pre-training our video models, following~\cite{li2022uniformerv2,wang2023videomaev2}. This results in a dataset with around 660K videos over 710 action classes, which is referred to as \textit{K710} in the following sections. 

\begin{table*}[]
    \centering
    \begin{subtable}[t]{.43\linewidth}
    \tablestyle{8pt}{1.0}
    \caption{Benefit of dynamic calibration. \textit{*}: w/o our init.}
    \centering
    \begin{tabular}{lccc}
    \toprule
      ~ & \bf \footnotesize Temporally & \bf\footnotesize SSV2 & \bf\footnotesize SSV2 \\
    \bf\footnotesize Calibration & \bf \footnotesize Varying  & \bf\footnotesize  Top-1 & \bf\footnotesize Top-1* \\
    \midrule
     None & \xmark & - & {32.0}\\
    \midrule
    \multirow{2}{*}{Learnable} & \xmark & 34.3 & 32.6 \\
    ~ & \cmark & 45.4 & 43.8 \\
    \midrule
    \multirow{2}{*}{Dynamic} & \xmark & 51.2 & 41.7 \\
    ~ & \cmark & 53.8 & 49.8 \\
    \midrule
    \graycell TAda & \graycell\cmark & \graycell\bf59.2 & \graycell\bf47.8 \\
    \bottomrule
    \end{tabular}
    \label{tab:calibrationsource}
    \caption{Calibration dimension.}
    \tablestyle{8pt}{1.0}
    \centering
    \begin{tabular}{lccc}
    \toprule
    \bf Cal. dim. & \bf $\mathbf{\Delta}_\text{Parms.}$  & \bf $\mathbf{\Delta}_\text{GFLOPs}$ &\bf Top-1 \\
    \midrule
    \bf\graycell$C_{\text{in}}$ & \bf\graycell 3.16M &\bf \graycell 0.016 &\bf\graycell 63.8 \\
    $C_{\text{out}}$ & 3.16M & 0.016 & 63.4 \\
    $C_{\text{in}}\times C_{\text{out}}$ & 4.10M & 0.024 & 63.7 \\
    $K^2$ & 2.24M & 0.009 & 62.7 \\
    \bottomrule
    \label{tab:calibrationdim}
    \end{tabular}
\end{subtable}
\begin{subtable}[t]{.55\linewidth}
    \caption{Plug-in evaluation of TAdaConv. }
    \centering
    \tablestyle{3pt}{1.0}
    \begin{tabular}{lcccll}
    \toprule
    & \bf\footnotesize TAda \\
    \bf \footnotesize Base Model & \bf \footnotesize Conv & \bf \footnotesize \#params. & \bf \footnotesize GFLOPs & \bf \footnotesize K400 & \bf \footnotesize SSV2\\
    \midrule
    \multirow{2}{*}{\cb \footnotesize SlowOnly {8\x8}$^\star$~\cite{slowfast}} & \xmark & 32.5M & 54.52 & 74.6 & 60.3 \\
    ~ & \cmark & \graycell 35.6M & \graycell 54.53 & \graycell 75.9 {\footnotesize (+1.3)} & \graycell 63.3 {\footnotesize (+3.0)}\\
    \midrule
    \multirow{2}{*}{\cb \footnotesize SlowFast 4\x16$^\star$~\cite{slowfast}} & \xmark & 34.5M & 36.10 & 75.0 & 56.7 \\
    ~ & \cmark & \graycell 37.7M & \graycell 36.11 & \graycell 76.5 {\footnotesize (+1.5)} & \graycell 59.8 {\footnotesize (+3.1)} \\
    \midrule
    \multirow{2}{*}{\cb \footnotesize SlowFast 8\x8$^\star$~\cite{slowfast}} & \xmark & 34.5M & 65.71 & 76.2 & 61.5 \\
    ~ & \cmark & \graycell 37.7M & \graycell 65.73 & \graycell 77.4 {\footnotesize (+1.2)}  & \graycell 63.9 {\footnotesize (+2.4)} \\
    \midrule
    \multirow{3}{*}{\cb \footnotesize R(2+1)D$^\star$~\cite{r21d}} & \xmark & 28.1M & 49.55 & 73.6 & 61.1 \\
    ~ & \cmark {\scriptsize 2d} & \graycell 31.2M & \graycell 49.57 & \graycell 75.2 {\footnotesize (+1.6)} & \graycell 62.9 {\footnotesize (+1.8)} \\
    ~ & \cmark {\scriptsize (2+1)d}& \graycell 34.4M & \graycell 49.58 & \graycell 75.4 {\footnotesize (+1.8)} & \graycell 63.8 {\footnotesize (+2.7)} \\
    \midrule
    \multirow{2}{*}{\cb \footnotesize R3D$^\star$~\cite{r21d}} & \xmark & 47.0M & 84.23 & 73.8 & 59.9 \\
    ~ & \cmark {\scriptsize 3d} & \graycell 50.1M & \graycell 84.24 & \graycell 74.9 {\footnotesize (+1.1)} & \graycell 62.9 {\footnotesize (+3.0)} \\
    \bottomrule
    \multicolumn{6}{l}{\makecell[l]{\scriptsize Notation $\star$ indicates our own implementation. }}\\
    \multicolumn{6}{l}{\makecell[l]{\scriptsize See \textit{Appendix} for details on the model structure. }}
    \label{tab:plugineval}
    \end{tabular}
\end{subtable}
\\
\begin{subtable}[t]{.48\linewidth}
    \tablestyle{12pt}{1.0}
    \caption{Calibration weight generation.
    \textit{K:} kernel size; \textit{Lin./Non-Lin.}: linear/non-linear weight generation; \textit{G:} global information $\mathbf{g}$.}
    \scalebox{0.9}{
    \begin{tabular}{lcccc}
    \toprule
    \bf Model & \bf TAdaConv & \bf K. & \bf G. & \bf Top-1\\
    \midrule
    TSN$^\star$ & - & - & - & 32.0 \\
    \midrule
    \multirow{9}{*}{Ours} & Lin. & 1 & \xmark & 37.5 \\
    ~ & Lin. & 3 & \xmark & 56.5 \\
    \cmidrule{2-5}
    ~ & Non-Lin. & (1, 1) & \xmark & 36.8\\
    ~ & Non-Lin. & (3, 1) & \xmark & 57.1\\
    ~ & Non-Lin. & (1, 3) & \xmark & 57.3\\
    ~ & Non-Lin. & (3, 3) & \xmark & 57.8\\
    \cmidrule{2-5}
    ~ & Lin. & 1 & \cmark & 53.4 \\
    ~ & Non-Lin. & (1, 1) & \cmark & 54.4\\
    ~ & \graycell\bf Non-Lin. & \graycell\bf(3, 3) & \graycell\cmark & \graycell\bf59.2\\
    \bottomrule
    \label{tab:calibrationweightgen}
    \end{tabular}}
\end{subtable}
\begin{subtable}[t]{.48\linewidth}
    \tablestyle{10pt}{1.0}
    \caption{Feature aggregation scheme.
    \textit{FA:} feature aggregation; \textit{Sc:} shortcut for convolution feature; \textit{SepBN:} separate batch norm.}
    \scalebox{0.9}{
    \begin{tabular}{cccccc}
    \toprule
    \bf TAdaConv & \bf FA. & \bf Sc. &\bf SepBN. & \bf Top-1 & $\mathbf{\Delta}$\\
    \midrule
    \xmark & - & - & - & 32.0 & - \\
    \cmark & - & - & - & 59.2 & +27.2\\
    \midrule
    \xmark & Avg. & \xmark & - & 47.9 & +15.9 \\
    \xmark & Avg. & \cmark & \xmark & 49.0 & +17.0 \\
    \xmark & Avg. & \cmark & \cmark & 57.0 & +25.0\\
    \midrule
    \cmark & Avg. & \xmark & - & 60.1 & +28.1 \\
    \cmark & Avg. & \cmark & \xmark & 61.5 & +29.5\\
    \graycell\cmark & \graycell \bf Avg. & \graycell\cmark & \graycell\cmark & \graycell\bf63.8 & \bf \graycell+31.8 \\
    \midrule
    \cmark & Max. & \cmark & \cmark & 63.5 & +31.5 \\
    \cmark & Mix. & \cmark & \cmark & 63.7 & +31.7\\
    \bottomrule
    \end{tabular}}
    \label{tab:featureaggregation}
\end{subtable}
    \vspace{-3mm}
    \caption{\textbf{Verification of hypothesis, plug-in evaluation, and in-depth ablative experiments on TAdaConv.} For plug-in evaluations, we plug TAdaConv into existing video recognition models and analyze the performance on both K400~\cite{kinetics400} and SSV2~\cite{ssv2}. For ablative experiments on TAdaConv, we mainly investigate its performance on SSV2. }
    \label{tab:ablative_experiments_tadaconv}
\end{table*}

\textbf{Training. }We train models initialized with ImageNet pre-training using AdamW~\cite{adamw} for 100/64/50 epochs on \textit{K400}, \textit{SSV1}/\textit{SSV2}, and \textit{EK100}, respectively. We adopt RandAugment~\cite{randaugment} for data augmentation and stochastic depth~\cite{huang2016droppath} and label smoothing~\cite{szegedy2016inceptionv3} for model regularization. 
We do not use Mixup~\cite{mixup} or Cutmix~\cite{cutmix} for both models. 
Exponential Moving Average (EMA)~\cite{polyak1992ema} is used for reducing overfitting during traning. For TAdaFormer with CLIP pre-trained weights~\cite{radford2021clip}, we shorten the schedule to 30/24/24 epochs respectively. See \textit{Appendix} for more details.

\subsection{Verification of hypothesis}
We start our experiments by verifying our hypothesis that \textit{relaxing the temporal invariance could lead to stronger temporal modeling capabilities of the video models.} To this end, we choose several sources for the calibration weights and compare the action classification performance on SSV2, with and without the relaxation of temporal invariance. The results are shown in Table~\ref{tab:calibrationsource}. 
It can be observed that both learnable and dynamic calibration can bring a notable improvement to the baseline with no calibration (TSN~\cite{tsn}), with dynamic calibration performing stronger than learnable calibration. On top of the calibrated models, making the weights vary along the temporal dimension can further boost classification accuracy, which means the model shows a better capability of temporal modeling when the temporal variance is relaxed.

\subsection{TAdaConv on existing video backbones}
TAdaConv is designed as a plug-in substitution for the spatial convolutions in the video models. 
As in Table~\ref{tab:plugineval}, TAdaConv improves the classification performance
with negligible computation overhead on a wide range of video models, including SlowFast~\cite{slowfast}, R3D~\cite{retrace} and R(2+1)D~\cite{r21d}, by an average of 1.3\% and 2.8\% respectively on K400 and SSV2 at an extra computational cost of less than 0.02 GFlops.
Further, not only can TAdaConv improve spatial convolutions, it also notably improve 3D and 1D convolutions.
For fair comparison, all models are trained using the same training strategy. 
Further plug-in evaluations for action classification is presented in \textit{Appendix}.
\subsection{Ablative anslysis on TAdaConv}
In this section, we thoroughly analyze our design choices and the effectiveness of TAdaConv and TAdaConvV2 in modeling temporal dynamics. We begin with TAdaConv, with SSV2 chosen as our main evaluation benchmark because of its more complex spatio-temporal relations. 

\textbf{Calibration weight initialization.} In Table~\ref{tab:calibrationsource}, we show that our initialization strategy for the calibration weight generation plays a critical role in dynamic weight calibration.
As in Table~\ref{tab:calibrationsource}, randomly initializing learnable weights slightly degrades the performance, while randomly initializing dynamic calibration weights (by randomly initializing the last layer of the weight generation function) notably degenerates the performance. 
It is likely that randomly initialized dynamic calibration weights perturb the pre-trained weights more severely than the learnable weights since it is dependent on the input.
Further comparisons on the initialization are shown in the Appendix.

\begin{figure}[t]
\centering
\includegraphics[width=\columnwidth]{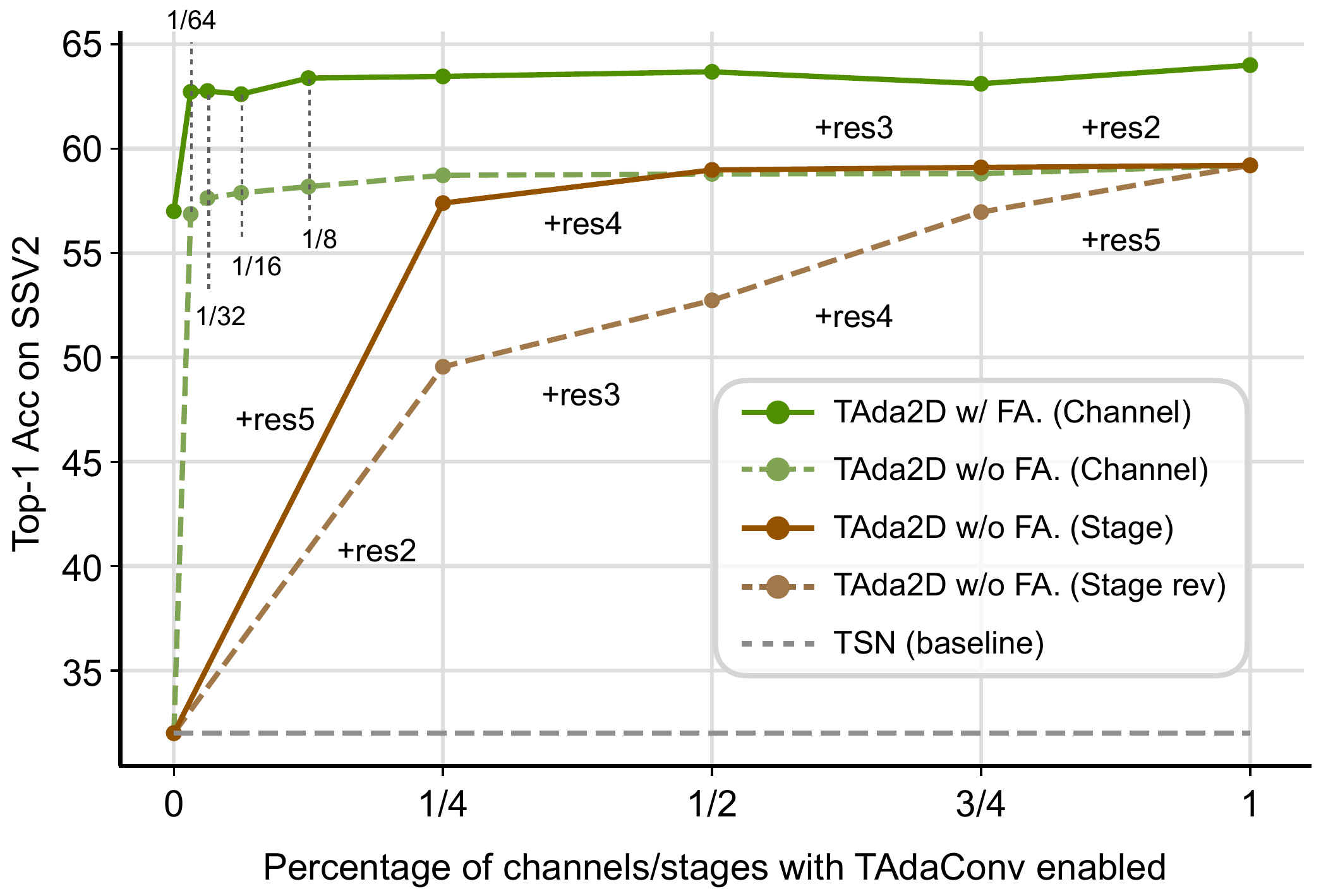}
\caption{
The classification accuracy of TAda2D on SSV2 with different channels (C.) and stages (S.) enabled.
}
\vspace{-3mm}
\label{fig:proportions}
\end{figure}

\textbf{Calibration weight generation function. } 
Having established that the temporally adaptive dynamic calibration with appropriate initialization can be an ideal strategy for temporal modeling, we further ablate different ways for generating the calibration weight in Table~\ref{tab:calibrationweightgen}. 
Linear weight generation function (\textit{Lin.}) applies a single 1D convolution to generate the calibration weight, while non-linear one (\textit{Non-Lin.}) uses two stacked 1D convolutions with batch normalizations and ReLU activation in between. 
When no temporal context is considered (K.=1 or (1,1)), TAdaConv can still improve the baseline but with a limited gap.
Enlarging the kernel size to cover the temporal context (K.=3, (1,3), (3,1) or (3,3)) effectively yields a boost of over 20\% on the accuracy, with K.=(3,3) having the strongest performance.
This shows the importance of the local temporal context during calibration weight generation.
Finally, for the scope of temporal context, introducing global context to frame descriptors performs similarly to only generating temporally adaptive calibration weights solely on the global context (in Table~\ref{tab:calibrationsource}). 
The combination of the global and temporal context yields a better performance for both variants.
In \textit{Appendix}, we also show 
that this function in our TAdaConv yields a better calibration on the base weight than existing dynamic filters.

\textbf{Feature aggregation. }
We ablate the aggregation scheme in TAda2D in Table~\ref{tab:featureaggregation}.
The performance is similar for plain aggregation $\mathbf{x}=\text{Avg}(\mathbf{x})$ and aggregation with a shortcut (Sc.) branch $\mathbf{x}=\mathbf{x}+\text{Avg}(\mathbf{x})$, with Sc. being slightly better. 
Separating the batchnorm (Eq.~\ref{eq:aggregation}) for the shortcut and the aggregation branch brings notable improvement.
Strided max and mix (avg+max) pooling slightly underperform the average pooling variant. 
Overall, the combination of TAdaConv and our feature aggregation scheme has an advantage over the TSN baseline of 31.8\%. 

\textbf{Calibration dimension. }
Multiple dimensions can be calibrated in the base weight.
Table~\ref{tab:calibrationdim} shows that calibrating the channel dimension more suitable than the spatial dimension, which means that the spatial structure of the original convolution kernel should be retained. 
Within channels, the calibration works better on $C_{\text{in}}$ than $C_{\text{out}}$ or both combined.
This is probably because the calibration weight generated by the input feature can better adapt to itself.

\textbf{Different stages employing TAdaConv. } Fig~\ref{fig:proportions} shows the stage by stage replacement of the spatial convolutions with TAdaConv in a ResNet.
A minimum improvement of 17.55\% is observed when TAdaConv is used in \textit{Res2}.
Compared to early stages, later stages contribute more to the final performance, as later stages provide more accurate calibration because of its rich semantics.
Overall, TAdaConv is used in all stages for the highest accuracy. 

\textbf{Different proportion of channels calibrated. }Here, we calibrate only a proportion of channels using TAdaConv and leave the other channels uncalibrated. The results are shown in Fig.~\ref{fig:proportions}. 
We find TAdaConv can improve the baseline by a large margin even if only 1/64 channels are calibrated, with larger proportion yielding further larger improvements.

\begin{table}[t]
\caption{Modernizing and improving TAdaBlocks. }
\vspace{-3mm}
\centering
\tablestyle{6pt}{1.0}
\begin{tabular}{llcc}
\toprule
\footnotesize Model & \footnotesize Variant & \footnotesize K400 & \footnotesize SSV2 \\
\midrule
ResNet2D & Baseline & 70.4 & 32.0 \\
$\ \ \ \ \ \ \ \downarrow$ & + TAdaConv & 73.9 & 59.2 \\
TAda2D & + T-Pool& 76.7 & 64.0 \\ 
\midrule
ConvNeXt & Baseline & 76.0 & 41.4 \\
{$\ \ \ \ \ \ \ \downarrow$} & + TAdaConv & 76.9 & 59.0 \\
TAdaConvNeXt-T & + T-Down & 78.4 & 64.8 \\
\multicolumn{1}{l}{\multirow{2}{*}{$\ \ \ \ \ \ \ \downarrow$}} & + TAdaConvV2 & 78.9 & 66.0 \\
 ~ & + T-Pool & 79.3 & 66.8\\
TAdaConvNeXtV2-T & + Stronger Aug & \graycell\bf 79.6 & \graycell\bf 67.2\\
\bottomrule
\end{tabular}
\label{tab:modernizing-tadablocks}
\end{table}

\subsection{Modernizing and improving TAdaBlocks}
We modernize our TAdaBlock following~\cite{convnext} and improve it with TAdaConvV2 and temporal aggregation in Table~\ref{tab:modernizing-tadablocks}. We observe a 5.6\% and 9.4\% improvement in the classification accuracy on K400 and SSV2, respectively, when we switch the base model from ResNet~\cite{resnet} to ConvNeXt~\cite{convnext}. Substituting the depth-wise convolution for TAdaConv further brings a 0.9\% and 17.6\% improvement. Following~\cite{arnab2021vivit,videoswin}, we employ a tubelet embedding stem (T-Down) in our TAdaConvNeXt, instantiated as a 3D convolution with temporal downsampling and an increased number of frames to keep the overall computation unchanged. 

On top of our TAdaConvNeXt model, we improve TAdaBlock by replacing TAdaConv with TAdaConvV2 and introducing the temporal aggregation scheme (T-Pool). The structural modification further leads to a performance gain of 0.9\% and 2.0\% on K400 and SSV2, respectively. Finally, with stronger augmentation (m7 to m9 for RandAugment~\cite{randaugment}), we achieve an accuracy of 79.6\% and 67.2\% on the two benchmarks with our tiny model. 

\subsection{Ablative analysis on TAdaConvV2 and TAdaBlocks}

\textbf{TAdaConvV2 and T-Pool in TAdaBlocks.} Table~\ref{tab:ablation-tadablocks} presents the ablative analysis on the TAdaBlock in both TAdaFormer and TAdaConvNeXtV2, specifically with respect to TAdaConvV2 and the temporal aggregation strategy. 

The baseline of TAdaFormer pretrained by CLIP~\cite{radford2021clip} demonstrates a strong spatial modeling capability, achieving an impressive accuracy of 83.6\% on K400. However, its ability to model complex dynamics is lacked. 
Introducing TAdaBlock with simple spatial convolution in between and no temporal aggregation brings negligible effect. 
On top of this, TAdaConvV2 notably improves the model in terms of temporal modeling, improving the performance on scene-related benchmark K400 by 0.9\% while bringing a 20\% performance gain on the temporal-related benchmark SSV2. On top of this, employing temporal aggregation (T-Pool) and tubelet embedding (Temp. Down.) further enhances the model's ability to model complex temporal dynamics. 

Compared to TAdaFormer, since TAdaConvNeXtV2 is pre-trained on ImageNet, the baseline performance is slightly lower. All three strategies bring notable improvements to both the scene- and temporal-centric benchmarks. 

\textbf{Pre-training. }We explore different pre-trained weights as initialization for TAdaConvNeXt and TAdaFormer in Table~\ref{tab:ablation-pretrain}. For TAdaConvNeXtV2, pre-training on K400 benefits SSV2 performance. For TAdaFormer, using pre-trained weights of CLIP~\cite{radford2021clip} outperforms the ImageNet pre-trained ones on both K400 and SSV2. 
CLIP+K710 initialization further improve the CLIP pre-trained variant by 2.1\% on K400, but the effect on SSV2 is less significant (0.1\%).
For the comparison against the state-of-the-art, we use ImageNet and CLIP as the default pre-training source respectively for TAdaConvNeXtV2 and TAdaFormer.

\begin{table}[t]
\caption{Ablation study on the TAdaBlock.}
\centering
\vspace{-3mm}
\tablestyle{2pt}{1.0}
\begin{tabular}{lccccc}
\toprule
~ & \multicolumn{2}{c}{\footnotesize{TAdaBlock}}& \footnotesize Temp. & ~ & ~ \\
\cline{2-3}
\footnotesize Model & \scriptsize TAdaConvV2 & \scriptsize T-Pool & \footnotesize Down. & \footnotesize K400 & \footnotesize SSV2 \\
\midrule
\vb \footnotesize{ViT-B/16} & \scriptsize N/A & \scriptsize N/A & \xmark & 83.6 & 48.1 \\
\vb\footnotesize{TAdaFormer-B/16} & \xmark & \xmark & \xmark & 83.6 & 48.2 \\
\vb\footnotesize{TAdaFormer-B/16} & \cmark & \xmark & \xmark & 84.5 & 68.6 \\
\vb\footnotesize{TAdaFormer-B/16} & \cmark & \cmark & \xmark & 84.5 & 69.2 \\
\vb\footnotesize{TAdaFormer-B/16} & \cmark & \cmark & \cmark & \graycell \bf 84.5 & \graycell \bf 70.4 \\
\midrule
\cb\footnotesize{ConvNeXt-T} & \xmark & \xmark & \xmark & 77.2 & 46.2 \\
\cb\footnotesize{TAdaConvNeXtV2-T} & \cmark & \xmark & \xmark & 78.0 & 63.3 \\
\cb\footnotesize{TAdaConvNeXtV2-T} & \cmark & \cmark & \xmark & 79.3 & 66.3\\
\cb\footnotesize{TAdaConvNeXtV2-T} & \cmark & \cmark & \cmark & \graycell \bf79.6 & \graycell \bf67.2 \\ 
\bottomrule
\end{tabular}
\label{tab:ablation-tadablocks}
\end{table}
\begin{table}[t]
\caption{Ablation study on different pre-training sources.}
\centering
\vspace{-3mm}
\tablestyle{6pt}{1.0}
\begin{tabular}{llcc}
\toprule
\footnotesize Model & \footnotesize Pretrain & \footnotesize K400 & \footnotesize SSV2 \\
\midrule
\multirow{2}{*}{TAdaConvNeXtV2-T} & IN1K & \graycell\bf79.6 & 65.2 \\
~ & IN1K+K400 & - & \graycell\bf67.2\\
\midrule
\multirow{4}{*}{TAdaFormer-B/16} & IN1K & 76.3 & 63.9 \\
~ & IN21K & 81.8 & 67.5 \\
~ & CLIP & \graycell 84.5 & \graycell 70.4 \\
~ & CLIP+K710 & 86.6 & 70.5 \\
\midrule
TimeSformer~\cite{timesformer} & IN21K & 78.7 & 59.5 \\
\midrule
\multirow{3}{*}{UniFormerV2-B/16~\cite{li2022uniformerv2}} & IN21K & 81.6 & 67.5 \\
~ & CLIP & 84.4 & 69.5 \\ 
~ & CLIP+K710 & 85.6 & - \\
\bottomrule
\end{tabular}
\label{tab:ablation-pretrain}
\vspace{-3mm}
\end{table}

\begin{table}[t]
\caption{Classification accuracy on Kinetics-400~\cite{kinetics400}. }
\centering
\vspace{-3mm}
\tablestyle{2pt}{1.1}
\begin{tabular}{lcrrc}
\toprule
\bf \footnotesize Model & \bf \footnotesize \#frames & \bf \footnotesize \#param. & \bf \scriptsize GFLOPs\x{views} & \bf \footnotesize Top-1 \\
\midrule
\multicolumn{5}{l}{\textit{\footnotesize Models without pretraining}}\\
\cb SlowFast 8\x8~\cite{slowfast} &  8+32 & 34.5M & 66\x3\x10 & 77.0 \\
\vb MViTv2-B~\cite{li2022mvitv2} & 32 & 51.2M & 225\x1\x5 & 82.9 \\
\midrule
\multicolumn{5}{l}{\textit{\footnotesize ImageNet-1K pretrained models}}\\
\cb TSM~\cite{tsm} & 8 & 24.3M & 43\x3\x10 & 74.1 \\
\cb TAda2D~\cite{huangtada} & 16 & 27.5M & 86\x3\x10 & 77.4  \\
\cb TAdaConvNeXt-T~\cite{huangtada} & 32 & 38.6M & 94\x3\x4 & 79.1  \\
\cb TANet~\cite{tam} & 16 & 25.6M & 242\x3\x4 & 79.3 \\
\cb TDN-R101~\cite{tdn} & 8+16 & - & 258\x3\x10 &  79.4 \\
\cb X3D-XXL~\cite{x3d} & - & 20.3M & 194\x3\x10 & 80.4 \\
\vb Swin-T~\cite{videoswin} & 32 & 28.2M & 88\x3\x4 & 78.8  \\
\vb Swin-S~\cite{videoswin} & 32 & 49.8M & 166\x3\x4 & 80.6 \\
\vb Swin-B~\cite{videoswin} & 32 & 88.1M & 282\x3\x4 & 80.6 \\
\cb MoViNet-A6~\cite{kondratyuk2021movinets} & 120 & 31.4M & 386\x1\x1 & 81.5 \\
\cb \graycell TAdaConvNeXtV2-T &\graycell 16 &\graycell 45.9M &\graycell 47\x3\x4 &\graycell 79.6 \\
\cb \graycell{TAdaConvNeXtV2-T} &\graycell32 &\graycell 45.9M &\graycell 94\x3\x4 &\graycell 80.8 \\
\cb \graycell{TAdaConvNeXtV2-S} &\graycell16 &\graycell 82.2M &\graycell 91\x3\x4 &\graycell 80.8 \\
\cb \graycell{TAdaConvNeXtV2-S} &\graycell32 &\graycell 82.2M &\graycell 183\x3\x4 &\graycell 81.9\\
\cb \graycell{TAdaConvNeXtV2-B} &\graycell 16 &\graycell 145.7M &\graycell 162\x3\x4 &\graycell 81.4 \\
\cb \graycell{TAdaConvNeXtV2-B} &\graycell 32 &\graycell 145.7M &\graycell 324\x3\x4 &\graycell 82.3 \\
\midrule
\multicolumn{5}{l}{\textit{\footnotesize ImageNet-21K pretrained models}}\\
\vb X-ViT~\cite{bulat2021xvit} & 16 & - & 283\x3\x1 & 80.2 \\
\vb TimeSformer~\cite{timesformer} & 96 & 121.4M & 2380\x3\x1 & 80.7  \\
\vb ViViT-L~\cite{arnab2021vivit} & 16 & 310.8M & 1446\x3\x4 & 80.6 \\
\vb MTV-B{\scriptsize$\uparrow320^2$}~\cite{yan2022multiview} & 32 & 310M & 930\x3\x4 & 82.4 \\
\vb Swin-B~\cite{videoswin} & 32 & 88.1M & 282\x3\x4 & 82.7 \\
\vb Swin-L~\cite{videoswin} & 32 & 197.0M & 604\x3\x4 & 83.1 \\
\vb MViT-v2-L{\scriptsize$\uparrow312^2$}~\cite{li2022mvitv2} & 40 & 217.6M & 2828\x3\x5 & 86.1 \\
\cb \graycell{TAdaConvNeXtV2-S} &\graycell32 &\graycell 82.2M &\graycell 183\x3\x4 &\graycell 82.9 \\
\cb \graycell{TAdaConvNeXtV2-B} &\graycell32 &\graycell 145.7M &\graycell 324\x3\x4 &\graycell 83.7 \\
\bottomrule
\end{tabular}
\label{tab:main-k400}
\vspace{-3mm}
\end{table}

\begin{table}[t]
\caption{
Classification accuracy on Kinetics-400~\cite{kinetics400} with large-scale pre-training and post-pre-training.
}
\centering
\tablestyle{1pt}{1.1}
\begin{tabular}{lcrrc}
\toprule
\bf \footnotesize Model & \bf \footnotesize \#frames & \bf \footnotesize \#param. & \bf \scriptsize GFLOPs\x{views} & \bf \footnotesize Top-1 \\
\hline
\multicolumn{5}{l}{\textit{\footnotesize Other large-scale pretrained models}}\\
\vb \gray MAE-ST~\cite{feichtenhofer2022maest} & \gray 16 & \gray 632M & \gray 1193\x3\x7 & \gray 85.1 \\
\vb \gray MAR~\cite{qing2023mar} & \gray 16 & \gray 311M & \gray 276\x3\x5 & \gray 85.3 \\
\vb \gray MaskFeat~\cite{wei2022maskfeat} & \gray 40 & \gray 218M & \gray 3790\x3\x4 & \gray 87.0 \\
\vb \gray CoVeR~\cite{zhang2021cover} {\footnotesize (JFT-3B)} & \gray 16 & \gray - & \gray - & \gray 87.2 \\
\vb \gray MTV-H{\footnotesize (WTS)}{\scriptsize$\uparrow280^2$}~\cite{yan2022multiview} & \gray 32 & - & \gray 6130\x3\x4 & \gray 89.9 \\
\vb \gray VideoMAE V2-g {\scriptsize$\uparrow266^2$}~\cite{wang2023videomaev2} & \gray 64 & \gray - & \gray 26716\x3\x2 & \gray 90.0 \\
\midrule
\multicolumn{5}{l}{\textit{\footnotesize CLIP pretrained models}}\\
\vb UniFormerV2-B/16~\cite{li2022uniformerv2} & 8 & 115M & {\scriptsize $\sim$}150\x3\x4 & 84.4 \\
\vb ST-Adapter-B/16~\cite{pan2022stadapter} & 32 & 93M & 607\x3\x1 & 82.0 \\
\vb EVL ViT-B/16~\cite{lin2022evl} & 32 & 115M & 592\x3\x1 & 84.2\\
\vb X-CLIP-B/16~\cite{ni2022xclip} & 16 & - & 287\x3\x4 & 84.7 \\
\vb ViFi-CLIP~\cite{rasheed2023vificlip} & 16 & 124.7M & 281\x4\x3 & 83.9 \\
\vb \graycell{TAdaFormer-B/16} &\graycell 16 &\graycell 104.1M &\graycell 153\x3\x4 &\graycell 84.5 \\
\vb ST-Adapter-L/14~\cite{pan2022stadapter} & 32 & 347M & 2749\x3\x1 & 87.2 \\
\vb EVL ViT-L/14~\cite{lin2022evl} & 32 & 363M & 2696\x3\x1 & 87.3\\
\vb X-CLIP-L/14~\cite{ni2022xclip} & 8 & - & 658\x3\x4 & 87.1 \\
\vb \graycell{TAdaFormer-L/14} &\graycell 16 &\graycell 364M &\graycell 703\x3\x4 &\graycell 87.6 \\
\midrule
\multicolumn{5}{l}{\textit{\footnotesize CLIP+K710 post-pretrained models}}\\
\vb UniFormerV2-B/16~\cite{li2022uniformerv2} & 8 & 115M & {\scriptsize $\sim$}150\x3\x4 & 85.6 \\
\cb \graycell{TAdaConvNeXtV2-S} & \graycell 32 & \graycell 82.2M & \graycell 183\x3\x4 & \graycell86.1 \\
\cb \graycell{TAdaConvNeXtV2-B} & \graycell 32 & \graycell 145.7M & \graycell 324\x3\x4 & \graycell86.4 \\
\vb \graycell{TAdaFormer-B/16} &\graycell 16 &\graycell 104.1M &\graycell 153\x3\x4 &\graycell 86.6 \\ 
\vb UniFormerV2-L/14~\cite{li2022uniformerv2} & 8 & 354M & {\scriptsize $\sim$}667\x3\x4 & 88.8 \\
\vb UniFormerV2-L/14~\cite{li2022uniformerv2} & 16 & 354M & {\scriptsize $\sim$}1334\x3\x4 & 89.1 \\
\vb UniFormerV2-L/14~\cite{li2022uniformerv2} & 32 & 354M & {\scriptsize $\sim$}2667\x3\x4 & 89.5 \\
\vb \graycell{TAdaFormer-L/14} &\graycell 16 &\graycell 364M &\graycell 703\x3\x4 &\graycell  88.9\\
\vb \graycell{TAdaFormer-L/14} &\graycell 32 &\graycell 364M &\graycell 1406\x3\x4 &\graycell  89.5\\
\vb \graycell{TAdaFormer-L/14} &\graycell 64 &\graycell 364M &\graycell 2812\x3\x4 &\graycell  89.9\\
\bottomrule
\end{tabular}
\label{tab:main-k400-pt}
\vspace{-3mm}
\end{table}
\begin{table}[t]
\caption{Classification accuracy on SSV1 and SSV2.
}
\centering
\tablestyle{2pt}{1.0}
\begin{tabular}{lcrrr}
\toprule
\footnotesize\bf Model & \footnotesize\bf \#frames & \bf \scriptsize GFLOPs\x{views}  & \footnotesize\bf SSV1 & \footnotesize\bf SSV2\\
\midrule
\cb TSM~\cite{tsm} & 16 & 86\x3\x2 & 47.2 & 63.4 \\
\cb MoViNet-A3~\cite{kondratyuk2021movinets} & 50 & 24\x1\x1 & - & 64.1 \\
\cb TANet~\cite{tam} & 16 & 86\x3\x2 & 47.6 & 64.6 \\
\cb TEANet~\cite{li2020tea} & 16 & 86\x1\x1 & 48.9 & - \\
\cb TEANet~\cite{li2020tea} & 16 & 86\x3\x10 & - & 65.1 \\
\cb TAda2D~\cite{huangtada} & 16 & 86\x3\x2 & - & 65.6 \\
\cb TAdaConvNeXt-T~\cite{huangtada} & 32 & 94\x3\x2 & - & 67.1 \\
\cb TDN-R101~\cite{tdn} & 8+16 & 258\x1\x1 & 56.8 &  68.2  \\
\cb \graycell{TAdaConvNeXtV2-T} &  \graycell 16  & \graycell 47\x3\x2 & \graycell 54.1 & \graycell 67.2 \\
\cb \graycell{TAdaConvNeXtV2-T} &  \graycell 32  & \graycell 94\x3\x2& \graycell 56.4 & \graycell 69.8 \\ 
\cb \graycell{TAdaConvNeXtV2-S} &\graycell16 &\graycell 91\x3\x2 & \graycell 55.6 &\graycell 68.4 \\
\cb \graycell{TAdaConvNeXtV2-S} &\graycell32 &\graycell183\x3\x2 & \graycell 58.5 &\graycell 70.0 \\ 
\cb \graycell{TAdaConvNeXtV2-S}$^\dagger$ &\graycell32 &\graycell183\x3\x2 & \graycell 59.7 &\graycell 70.6 \\ 
\cb \graycell{TAdaConvNeXtV2-B}$^\dagger$ &\graycell32 &\graycell 324\x3\x2 & \graycell 60.7 &\graycell 71.1 \\ 
\midrule
\vb ViViT-L/16x2 FE~\cite{arnab2021vivit} & 32 & 903\x3\x4 & - & 65.4 \\
\vb X-ViT~\cite{bulat2021xvit} & 16 & 283\x3\x1 & - & 67.2 \\
\vb MTV-B{\scriptsize$\uparrow320^2$}~\cite{yan2022multiview} & 32 & 930\x3\x4 & -& 68.5 \\
\vb Swin-B$^\dagger$\cite{videoswin} & 32 & 321\x3\x1 & - & 69.6 \\
\vb MViTv2-B~\cite{li2022mvitv2} & 32 & 225\x3\x1 & - & 70.5 \\
\vb ST-Adapter-B/16$^{\star}$~\cite{pan2022stadapter} & 32 & 651\x3\x1 & - & 69.5 \\
\vb ST-Adapter-L/14$^{\star}$~\cite{pan2022stadapter} & 32 & 2749\x3\x1 & - & 72.3 \\
\vb UniFormerV2-B/16$^{\star}$~\cite{li2022uniformerv2} & 32 & {\scriptsize $\sim$}370\x3\x2 & 59.5 & 71.0 \\
\vb UniFormerV2-L/14$^{\star}$~\cite{li2022uniformerv2} & 32 & {\scriptsize $\sim$}1716\x3\x2 & 62.9 & 73.1 \\
\vb MViTv2-L{\scriptsize$\uparrow312^2$}~\cite{li2022mvitv2} & 40 & 2828\x3\x1 & - & 73.3 \\
\vb \graycell{TAdaFormer-B/16}$^{\star}$ & \graycell 16 & 187\x3\x2\graycell & \graycell 59.2 & \graycell 70.4 \\
\vb \graycell{TAdaFormer-B/16}$^{\star}$ & \graycell 32 & \graycell 374\x3\x2& \graycell 61.2 & \graycell 71.3 \\
\vb \graycell{TAdaFormer-L/14}$^{\star}$ & \graycell 16 & \graycell 858\x3\x2 & \graycell 62.0 & \graycell 72.4 \\
\vb \graycell{TAdaFormer-L/14}$^{\star}$ & \graycell 32 & \graycell 1716\x3\x2 & \graycell 63.7 & \graycell 73.6 \\
\bottomrule
\multicolumn{5}{l}{\footnotesize $^\dagger$ indicates initialization with ImageNet21K+K400 pre-training.}\\
\multicolumn{5}{l}{\footnotesize $^{\star}$ indicates initialization with CLIP-400M pre-training.}\\
\end{tabular}
\label{tab:main-ssv2}
\end{table}

\subsection{Main results}
\begin{table}[t]
\caption{
Classification accuracy on Epic-Kitchens-100~\cite{ek100}. $\bm{\uparrow}$ indicates the main evaluation metric for the dataset.}
\centering
\tablestyle{5.5pt}{1.0}
\begin{tabular}{lccc}
\toprule
\bf Model & \textbf{Act.}$\bm{\uparrow}$ & Verb & Noun \\
\midrule
\cb TSN~\cite{tsn} & 33.2 & 60.2 & 46.0 \\
\cb TRN~\cite{trn} & 35.3 & 65.9 & 45.4 \\ 
\cb TSM~\cite{tsm} & 38.3 & 67.9 & 49.0 \\
\cb SlowFast~\cite{slowfast} & 38.5 & 65.6 & 50.0 \\
\cb TAda2D~\cite{huangtada} & 41.6 & 65.1 & 52.4\\
\cb ir-CSN-152~\cite{vivitek100} & 44.5 & 68.4 & 55.9 \\
\cb MoViNet-A6~\cite{kondratyuk2021movinets} & 47.7 & 72.2 & 57.3 \\
\cb\graycell TAdaConvNeXtV2-T {\footnotesize(IN1K)} &\graycell 42.4 &\graycell 67.1 &\graycell 53.7 \\
\cb\graycell TAdaConvNeXtV2-T {\footnotesize(K710)} &\graycell 47.4 &\graycell 70.4 &\graycell 58.6 \\
\cb\graycell TAdaConvNeXtV2-S {\footnotesize(K710)} &\graycell 48.9 &\graycell 71.0 &\graycell 60.2 \\
\midrule
\vb ViViT-L/16x2 FE~\cite{arnab2021vivit} & 44.0 & 66.4 & 56.8 \\
\vb X-ViT~\cite{bulat2021xvit} & 44.3 & 68.7 & 56.4 \\
\vb ViViT-B/16x2 FE {\scriptsize$\uparrow384$}~\cite{vivitek100} & 47.0 & 67.2 & 59.0 \\
\vb ST-Adapter-B/16~\cite{pan2022stadapter} & - & 67.6 & 55.0 \\
\vb MeMViT~\cite{wu2022memvit} & 48.4 & 71.4 & 60.3 \\
\vb MTV-B{\scriptsize$\uparrow320^2$}~\cite{yan2022multiview} & 48.6 & 68.0 & 63.1 \\
\vb MTV-B{\footnotesize(WTS)}{\scriptsize$\uparrow280^2$}~\cite{yan2022multiview} & 50.5 & 69.9 & 63.9 \\
\graycell\vb TAdaFormer-B/16 {\footnotesize(K710)} & \graycell49.1 & \graycell71.0 & \graycell60.5 \\
\graycell\vb TAdaFormer-L/14 {\footnotesize(K710)} & \graycell51.8 & \graycell71.7 & \graycell64.1 \\
\bottomrule
\end{tabular}
\label{tab:main-ek100}
\end{table}

\begin{table}[t]
\caption{
Zero-shot classification on UCF101~\cite{soomro2012ucf101} and HMDB51~\cite{kuehne2011hmdb}.}
\centering
\tablestyle{5.5pt}{1.0}
\begin{tabular}{lcc}
\toprule
\bf Model & HMDB-51 & UCF-101 \\
\midrule
\nb MTE~\cite{xu2016mte} & 19.7 $\pm$ 1.6 & 15.8 $\pm$ 1.3 \\
\nb ASR~\cite{wang2017asr} & 21.8 $\pm$ 0.9 & 24.4 $\pm$ 1.0 \\
\nb ER-ZSAR~\cite{chen2021er-zsar} & 35.3 $\pm$ 4.6 & 51.8 $\pm$ 2.9 \\
\midrule
\vb CLIP~\cite{radford2021clip} & 40.8 $\pm$ 0.3 & 63.2 $\pm$ 0.2 \\
\vb ActionCLIP~\cite{wang2021actionclip} & 40.8 $\pm$ 5.4 & 58.3 $\pm$ 3.4\\
\vb X-CLIP-B/16~\cite{ni2022xclip} & 44.6 $\pm$ 5.2 & 72.0 $\pm$ 2.3 \\
\vb A5~\cite{ju2022a5} & 44.3 $\pm$ 2.2 & 69.3 $\pm$ 4.2 \\
\vb ViFi-CLIP~\cite{rasheed2023vificlip} & 51.3 $\pm$ 
 0.6 & 76.8 $\pm$ 0.7 \\
\graycell\vb TAdaFormer-B/16 & \graycell52.1 $\pm$ 1.4 & \graycell78.5 $\pm$ 1.2\\
\graycell\vb TAdaFormer-L/14 & \graycell57.2 $\pm$ 0.7 & \graycell81.1 $\pm$ 0.9\\
\midrule
\graycell\vb TAdaFormer-B/16 {\footnotesize(K710)} & \graycell55.9 $\pm$ 0.4 & \graycell79.5 $\pm$ 0.7\\
\graycell\vb TAdaFormer-L/14 {\footnotesize(K710)} & \graycell59.7 $\pm$ 0.5 & \graycell83.0 $\pm$ 0.7\\
\bottomrule
\end{tabular}
\label{tab:main-zeroshot}
\end{table}
\begin{table}[t]
    \centering
    \caption{Action localization on HACS~\cite{hacs}.}
    \tablestyle{4.5pt}{1.0}
    \begin{tabular}{lcccccc}
        \toprule
        ~ & \multicolumn{6}{c}{\bf HACS} \\
        \cmidrule(r){2-7}
        Model & @0.5 & @0.6 & @0.7 & @0.8 & @0.9 & \bf Avg.$\bm{\uparrow}$ \\
        \midrule
        SSN~\cite{zhao2017ssn} & 28.8 & - & - & - & - & 19.0\\
        G-TAD~\cite{xu2020gtad} & 41.1 & - & - & - & - & 27.5\\
        TadTR~\cite{liu2022tadtr} & 47.1 & - & - & - & - & 32.1 \\
        \midrule
        \textit{\scriptsize BMN~\cite{bmn}+}\\
        TSN~\cite{huangtada} & 43.6 & 37.7 & 31.9 & 24.6 & 15.0 & 28.6 \\
        TAda2D~\cite{huangtada} & 48.7 & 42.7 & 36.2 & 28.1 & 17.3 & 32.3 \\
        \graycell TAdaFormer-L/14 & \graycell51.3 & \graycell44.8 & \graycell38.0 & \graycell30.0 & \graycell18.6 & \graycell34.1 \\
        \graycell TAdaConvNeXt-S &  \graycell53.3 & \graycell47.0 & \graycell40.2 & \graycell32.0 & \graycell20.2 & \graycell36.1\\
        \bottomrule
    \end{tabular}
    \label{tab:hacs-localization}
\end{table}
\begin{table}[t]
    \caption{Action localization on Epic-Kitchens-100~\cite{ek100}.}
    \tablestyle{2pt}{1.0}
    \begin{tabular}{llcccccc}
        \toprule
        ~ & \multicolumn{7}{c}{\bf Epic-Kitchens-100} \\
        \cmidrule(r){2-8}
        Model & Task & @0.1 & @0.2 & @0.3 & @0.4 & @0.5 & \bf Avg.$\bm{\uparrow}$ \\
        \midrule
        \multirowcell{3}[0pt][l]{BMN~\cite{bmn}\\+TSN} & Verb & 15.98 & 15.01 & 14.09 & 12.25 & 10.01 & 13.47\\
        ~ & Noun &15.11 & 14.15 & 12.78 & 10.94 & 8.89 & 12.37\\
        ~ & \bf Act.$\bm{\uparrow}$ & 10.24 & 9.61 & 8.94 & 7.96 & 6.79 & 8.71\\
        \midrule
        \multirowcell{3}[0pt][l]{BMN~\cite{bmn}\\+TAda2D~\cite{huangtada}} & Verb & 19.70 & 18.49 & 17.41 & 15.50 & 12.78 & 16.78 \\
        ~ & Noun & 20.54 & 19.32 & 17.94 & 15.77 & 13.39 & 17.39 \\
        ~ & \bf Act.$\bm{\uparrow}$ & 15.15 & 14.32 & 13.59 & 12.18 & 10.65 & 13.18 \\
        \midrule
        \multirowcell{3}[0pt][l]{BMN~\cite{bmn}\\+TAdaFormer-L/14} & Verb & 20.87 & 20.09 & 18.99 & 16.42 & 13.81 & 18.03 \\
        ~ & Noun & 27.75 & 26.28 & 24.51 & 21.86 & 17.97 & 23.67 \\
        ~ & \bf Act.$\bm{\uparrow}$ & 20.39 & 19.35 & 18.28 & 16.35 & 14.51 & 17.85 \\
        \midrule
        \multirowcell{3}[0pt][l]{BMN~\cite{bmn}\\+TAdaConvNeXt-S} & Verb &17.81 & 16.94 & 16.05 & 14.25 & 11.89 & 15.39 \\
        ~ & Noun &21.90 & 20.92 & 19.33 & 17.22 & 14.68 & 18.81 \\
        ~ & \bf Act.$\bm{\uparrow}$ &15.61 & 14.80 & 13.73 & 12.35 & 10.90 & 13.47 \\
        \toprule
        \multirowcell{3}[0pt][l]{ActionFormer~\cite{zhang2022actionformer}\\+SlowFast} & Verb & 26.58 & 25.42 & 24.15 & 22.29 & 19.09 & 23.51\\
        ~ & Noun & 25.21 & 24.11 & 22.66 & 20.47 & 16.97 & 21.88\\
        ~ & \bf Act.$\bm{\uparrow}$ & 18.40 & 17.71 & 16.80 & 15.65 & 13.52 & 16.42\\
        \midrule
        \multirowcell{3}[0pt][l]{ActionFormer~\cite{actionformer-ek100-2022-challenge-report}\\+SlowFast\&ViViT} & Verb & 26.97 & 25.90 & 24.21 & 21.77 & 18.47 & 23.46\\
        ~ & Noun & 28.61 & 27.14 & 24.92 & 22.13 & 18.69 & 24.30 \\ 
        ~ & \bf Act.$\bm{\uparrow}$ & 23.90 & 22.98 & 21.37 & 19.57 & 16.94 & 20.95 \\
        \midrule
        \multirowcell{3}[0pt][l]{ActionFormer\\+TAdaConvNeXt-S} & Verb & 29.11 & 28.37 & 26.99 & 24.22 & 20.64 & 25.86 \\
        ~ & Noun & 29.21 & 27.94 & 26.22 & 23.54 & 18.73 & 25.13 \\
        ~ & \graycell\bf Act.$\bm{\uparrow}$ & \graycell20.78 & \graycell19.75 & \graycell18.56 & \graycell17.07 & \graycell14.54 & \graycell18.14 \\
        \midrule
        \multirowcell{3}[0pt][l]{ActionFormer\\+TAdaFormer-L/14} & Verb & 32.08 & 31.09 & 29.40 & 26.64 & 22.71 & 28.38\\
        ~ & Noun & 35.00 & 33.42 & 30.98 & 27.32 & 22.36 & 29.82 \\
        ~ & \graycell\bf Act.$\bm{\uparrow}$ & \graycell24.92 & \graycell23.68 & \graycell22.33& \graycell20.61 & \graycell18.29 & \graycell\bf21.97 \\
        \bottomrule
    \end{tabular}
    \label{tab:ek100-localization}
\end{table}
\textbf{Kinetics-400. }Table~\ref{tab:main-k400} shows the results on Kinetics-400 without large-scale pre-training. TAdaConvNeXtV2 surpasses most existing approaches with a similar computation budget both when pre-trained on ImageNet-1K and ImageNet-21K. A highlight is observed where our TAdaConvNeXtV2-S with 32 frames outperforms Swin-B by 1.3 using only 57\% of the computation. 

Table~\ref{tab:main-k400-pt} presents the comparison for models with large-scale pre-training. Compared to existing CLIP pre-trained models, TAdaFormer achieves competitive performance. When post-pre-trained on K710, TAdaFormer outperforms UniFormerV2 by a notable margin under similar computation budgets. We also observe better scalability of TAdaFormer when it is compared with TAdaConvNeXtV2. 

\textbf{Something-Something-V1 and V2. }We show the performance comparison on temporal-related datasets, \textit{i.e.,} SSV1 and SSV2, in Table~\ref{tab:main-ssv2}. TAdaConvNeXt and TAdaFormer achieve a favorable performance against existing convolutional and transformer-based models with identical or similar pre-training sources, respectively. Compared to the best convolutional model TDN-R101, TAdaConvNeXt-B outperforms it by 3.9 and 2.9 on SSV1 and SSV2. Compared to CLIP-pre-trained UniFormerV2-L/14, TAdaFormer-L/14 achieves an improvement of 0.8 and 0.5 on the two datasets. 

\textbf{Epic-Kitchens-100.} We compare the performance on ego-centric action recognition in Table~\ref{tab:main-ek100}. Compared to existing convolutional models, our TAdaConvNeXtV2-S achieves a favourable performance. Notably, we observe a higher accuracy for TAdaConvNeXt models on noun recognition in ego-centric videos. Transformer-based models are generally stronger than convolutional ones on EK100, where our TAdaFormer achieves a competitive performance with existing Transformers for video understanding. 

\textbf{Zero-shot classification on UCF101 and HMDB51.} To more comprehensively evaluate our TAdaFormer, we include the results on zero-shot classification in Table~\ref{tab:main-zeroshot}. Here, we initialize the model with CLIP pre-trained weights and train our TAdaFormer with the corresponding language model~\cite{radford2021clip}. We observe a notable improvement of TAdaFormer-B/16 on both datasets compared to the fine-tuned CLIP ViFi-CLIP~\cite{rasheed2023vificlip}. 
On top of this, we find scaling up the model and pre-training brings a further boost to the zero-shot performance.

\section{Evaluations on action localization}
\label{sec:exp-tal}

\textbf{Dataset, pipeline, and evaluation.} 
Action localization is an essential task for understanding untrimmed videos, whose current pipeline makes it heavily dependent on the quality of the video representations.
We evaluate our TAdaConvNeXtV2 and TAdaFormer on two large-scale action localization datasets, HACS~\cite{hacs} and Epic-Kitchens-100~\cite{ek100}.
The general pipeline follows~\cite{ek100,ek100actionlocalization,hacscompetition}, which uses Boundary Matching Network (BMN)~\cite{bmn} for generating action boundaries.
For evaluation, we use the average mean Average Precision (average mAP) at IoU [0.5:0.05:0.95] for HACS and [0.1:0.1:0.5] for EK100, following the standard protocol.
More details are included in the \textit{Appendix.}

\textbf{Main results.} 
We present the results on the two datasets in Table~\ref{tab:hacs-localization} and Table~\ref{tab:ek100-localization}. On HACS, we found BMN~\cite{bmn} using TAdaFormer and TAdaConvNeXt features yields a favourable performance compared to some recent methods. On Epic-Kitchens-100, we further employ ActionFormer~\cite{zhang2022actionformer} and found TAdaFormer stronger than the ensemble of ViViT and SlowFast. Overall, we found TAdaConvNeXt and TAdaFormer provide strong features for localzing actions in long videos.

\section{Conclusions}
Based on our preliminary work~\cite{huangtada}, this work presents TAdaConvV2 in replacement of the convolution operations in existing models for video understanding, and two strong video models, \textit{i.e.,} TAdaConvNeXtV2 and TAdaFormer. With large-scale pre-training and post-pre-training, our video models demonstrate competitive performances to the state-of-the-art approaches, both in the task of action recognition and localization. We hope our work can facilitate further research in video understanding.

\ifCLASSOPTIONcompsoc
  \section*{Acknowledgments}
\else
  \section*{Acknowledgment}
\fi
This research is supported by the Agency for Science, Technology and Research (A*STAR) under its AME Programmatic Funding Scheme (Project \#A18A2b0046), by the RIE2020 Industry Alignment Fund – Industry Collaboration Projects (IAF-ICP) Funding Initiative, as well as cash and in-kind contribution from the industry partner(s), and by Alibaba Group through Alibaba Research Intern Program. 


\ifCLASSOPTIONcaptionsoff
  \newpage
\fi



%


\bibliographystyle{IEEEtran}
\bibliography{citation}

\appendices
\section{Overview}
\renewcommand{\thetable}{A\arabic{table}}
\renewcommand{\thefigure}{A\arabic{figure}}
\setcounter{figure}{0}
\setcounter{table}{0}
In the appendix, we provide detailed analysis on the temporal convolutions (Appendix~\ref{appendix:tempconvs}), further implementation details (Appendix~\ref{appendix:implementationdetails}) on the action classification and localization, model structures that we used for evaluation (Appendix~\ref{appendix:modelstructure}), per-category improvement analysis on Something-Something-V2 (Appendix~\ref{appendix:percategoryimprovement}), further plug-in evaluations on Epic-Kitchens classification (Appendix~\ref{appendix:pluginclassification}) plug-in evaluations on the temporal action localization task (Appendix~\ref{appendix:pluginlocalization}), the visualization of the training procedure of TSN and TAda2D (Appendix~\ref{appendix:trainingprocedure}), as well as detailed comparisons between TAdaConv and existing dynamic filters (Appendix~\ref{appendix:comparisondynamicfilter}).

\section{Detailed analysis on temporal convolutions}
\label{appendix:tempconvs}
Here, we provide a detailed analysis to showcase the underlying process of temporal modeling by temporal convolutions. As in Sec.~{\color{forestgreen}{3.1}}, we use depth-wise temporal convolutions for simplicity and its wide application. We first analyze the case where temporal convolutions are directly placed after spatial convolutions without non-linear activation in between, before activation functions are inserted in the second part of our analysis.

\textbf{Without activation. }We first consider a simple case with no non-linear activation functions between the temporal convolution and the spatial convolution. 
Given a 3\x1\x1 depth-wise temporal convolution parameterized by $\bm{\beta}=[\bm{\beta}_1, \bm{\beta}_2, \bm{\beta}_3]$, where $\bm{\beta}_1, \bm{\beta}_2, \bm{\beta}_3 \in \mathbb{R}^{C_o}$, a spatial convolution parameterized by $\mathbf{W}\in\mathbb{R}^{C_o\times C_i\times K^2}$, the output feature $\mathbf{\tilde{x}_t}$ of the $t$-th frame can be obtained by:
\begin{equation}
    \mathbf{\tilde{x}}_t = \bm{\beta}_1 \cdot (\mathbf{W} * \mathbf{x}_{t-1}) + \bm{\beta}_2 \cdot (\mathbf{W} * \mathbf{x}_{t}) + \bm{\beta}_3 \cdot (\mathbf{W} * \mathbf{x}_{t+1})\ ,
\end{equation}
\noindent where $\cdot$ denotes element-wise multiplication with broadcasting, and $*$ denotes convolution over the spatial dimension. 
In this case, $\bm{\beta}$ could be grouped with the spatial convolution weight $\mathbf{W}$ and the combination of temporal and spatial convolution can be rewritten as:
\begin{equation}
    \mathbf{\tilde{x}}_t =\mathbf{W}_{t-1} * \mathbf{x}_{t-1} + \mathbf{W}_{t} * \mathbf{x}_{t} + \mathbf{W}_{t+1} * \mathbf{x}_{t+1}\ ,
\end{equation}
\noindent where $\mathbf{W}_{t-1}=\bm{\beta}_1\cdot\mathbf{W}$, $\mathbf{W}_{t}=\bm{\beta}_2\cdot\mathbf{W}$ and $\mathbf{W}_{t+1}=\bm{\beta}_3\cdot\mathbf{W}$. 
This equation shares the same form with the Eq.~{\color{forestgreen}2} in the manuscript.
In this case, the combination of temporal convolution with spatial convolution can be certainly viewed as the temporal convolution simply performs calibration on spatial convolutions before aggregation, with different weights assigned to different time steps for the calibration. 

\textbf{With activation.} Next, we consider a case where activation is in between the temporal convolution and spatial convolution. The output feature $\mathbf{\tilde{x}}_t$ are now obtained by:
\begin{equation}
    \mathbf{\tilde{x}}_t = \bm{\beta}_1 \cdot \delta(\mathbf{W} * \mathbf{x}_{t-1}) + \bm{\beta}_2 \cdot \delta(\mathbf{W} * \mathbf{x}_{t}) + \bm{\beta}_3 \cdot \delta(\mathbf{W} * \mathbf{x}_{t+1})\ .
\end{equation}

Next, we show that this can be still rewritten in the form of Eq.~{\color{forestgreen}2}. Here, we consider the case where ReLU~\cite{relu} is used as the activation function, denoted as $\delta$:
\begin{equation}
    \delta(x) = \left\{\begin{aligned}
        &x \quad &x > 0 \\
        &0 \quad &x\leq 0
    \end{aligned}\right. \ .
\end{equation}
Hence, the term $\delta(\mathbf{W}*\mathbf{x}_t)$ can be easily expressed as:
\begin{equation}
    \delta(\mathbf{W}*\mathbf{x}_t) = \mathbf{M}_t\cdot \mathbf{W}*\mathbf{x}_t\ ,
\end{equation}
\noindent where $\mathbf{M}_t\in\mathbb{R}^{C\times H\times W}$ is a binary map sharing the same shape as $\mathbf{x}_t$, indicating whether the corresponding element in $\mathbf{W}*\mathbf{x}_t$ is greater than 0 or not. That is:
\begin{equation}
    \mathbf{M}_t^{(c,i,j)} = \left\{\begin{aligned}
        &1 \quad &\text{if} \quad &(\mathbf{W}*\mathbf{x}_t)^{(c,i,j)} > 0\\
        &0 \quad &\text{if} \quad &(\mathbf{W}*\mathbf{x}_t)^{(c,i,j)}\leq 0
    \end{aligned}\right. \ ,
\end{equation}
\noindent where $c, i, j$ are the location index in the tensor.
Hence, with activation, temporal convolution can be expressed as:
\begin{equation}
    \mathbf{\tilde{x}}_t = \bm{\beta}_1 \cdot \mathbf{M}_{t-1} \cdot \mathbf{W} * \mathbf{x}_{t-1} + \bm{\beta}_2 \cdot \mathbf{M}_{t} \cdot \mathbf{W} * \mathbf{x}_{t} + \bm{\beta}_3 \cdot \mathbf{M}_{t+1} \cdot \mathbf{W} * \mathbf{x}_{t+1}\ .
\end{equation}
\noindent In this case, we can set $\mathbf{W}_{t-1}^{(i,j)}=\bm{\beta}_1\cdot\mathbf{M}_{t-1}^{(i,j)}\cdot\mathbf{W}$, $\mathbf{W}_{t}^{(i,j)}=\bm{\beta}_2\cdot\mathbf{M}_{t}^{(i,j)}\cdot\mathbf{W}$, and $\mathbf{W}_{t+1}^{(i,j)}=\bm{\beta}_3\cdot\mathbf{M}_{t+1}^{(i,j)}\cdot\mathbf{W}$, where $(i,j)$ indicate the spatial location index.
In this case, each filter for a specific time step $t$ is composed of $H\times W$ filters and Eq.~{\color{forestgreen}1} can be rewritten as Eq.~{\color{forestgreen}2}. 
Interestingly, it can be observed that with ReLU activation function, the convolution weights are different for all spatio-temporal locations, since the binary map $\mathbf{M}$ depends on the results of the spatial convolutions.

\section{Further implementation details}
\label{appendix:implementationdetails}
Here, we further describe the implementation details for the action classification and action localization experiments.
For fair comparisons, we keep all the training strategies the same for our baseline, the plug-in evaluations as well as our own models.

\begin{table*}[t]
    \centering
    \tablestyle{10pt}{1.0}
    \begin{tabular}{l|ccccc}
         \toprule
         training config & K710 & K400 (K710) & K400 (ImageNet) & {SSV1/SSV2} & EK100\\
         \midrule
         optimizer & \multicolumn{5}{c}{AdamW~\cite{adamw}} \\
         learning rate schedule & \multicolumn{5}{c}{cosine decay} \\
         weight decay & \multicolumn{5}{c}{0.02} \\
         optimizer momentum & \multicolumn{5}{c}{$\beta_1,\beta_2=0.9,0.999$} \\
         dropout~\cite{dropout} & \multicolumn{5}{c}{0.5} \\
         clip grading & \multicolumn{5}{c}{None} \\
         \midrule
         base learning rate & \multicolumn{5}{c}{5e-4}\\
         batch size & \multicolumn{5}{c}{512} \\
         training epochs & 100 & 30 & 100 & 64 & 50\\
         warmup epochs & 8 & 4 & 8 & 2.5 & 5\\
         randaugment~\cite{randaugment} & \multicolumn{5}{c}{(9, 0.5)}\\
         label smoothing~\cite{szegedy2016inceptionv3} & \multicolumn{4}{c}{0.0} & 0.1 \\
         \multirow{3}{*}{stochastic depth} & 0.2 (T) & 0.2 (T) & 0.2 (T) & 0.3 (T) & 0.3 (T)\\
         ~ & 0.4 (S) & 0.4 (S) & 0.4 (S) & 0.5 (S) & 0.5 (S)\\
         ~ & 0.6 (B) & 0.6 (B) & 0.6 (B) & 0.6 (B) & -\\
         \bottomrule
    \end{tabular}
    \caption{TAdaConvNeXtV2 training settings on K710, K400, SSV1/SSV2, and EK100.}
    \label{tab:training_tadaconvnextv2}
\end{table*}

\subsection{Action classification with TAdaConvNeXtV2}
We evaluate our approach on action classification using four large-scale benchmarks. We list the training configurations for TAdaConvNeXtV2 and TAdaFormer on action classification benchmarks in Table~\ref{tab:training_tadaconvnextv2} and Table~\ref{tab:training_tadaformer}, respectively. 
\begin{table*}[t]
    \centering
    \tablestyle{10pt}{1.0}
    \begin{tabular}{l|cc|cc|cc|cc|cc}
        \toprule
         training config & \multicolumn{2}{c}{K710} & \multicolumn{2}{c}{K400 (K710)} & \multicolumn{2}{c}{K400 (CLIP)} & \multicolumn{2}{c}{SSV1/SSV2} & \multicolumn{2}{c}{EK100} \\
         \midrule
         optimizer& \multicolumn{10}{c}{AdamW~\cite{adamw}} \\
         learning rate schedule & \multicolumn{10}{c}{cosine decay} \\
         weight decay & \multicolumn{10}{c}{0.05} \\
         optimizer momentum & \multicolumn{10}{c}{$\beta_1,\beta_2=0.9,0.999$} \\
         dropout~\cite{dropout} & \multicolumn{10}{c}{0.5} \\
         clip grading & \multicolumn{10}{c}{None} \\
         EMA~\cite{polyak1992ema} & \multicolumn{10}{c}{0.9996} \\
         \midrule
         ~ & \scriptsize\textit{Base} & \scriptsize\textit{Large} & \scriptsize\textit{Base} & \scriptsize\textit{Large} & \scriptsize\textit{Base} & \scriptsize\textit{Large} & \scriptsize\textit{Base} & \scriptsize\textit{Large} & \scriptsize\textit{Base} & \scriptsize\textit{Large}\\
         \cmidrule(lr){2-3} \cmidrule(lr){4-5} \cmidrule(lr){6-7} \cmidrule(lr){8-9} \cmidrule(lr){10-11}
         base learning rate & 1e-4 & 5e-5  & 1e-5  & 5e-6 & 5e-5 & 2e-5 & 5e-4 & 2.5e-4 & 2.5e-4 & 1e-4 \\
         batch size & 512  & 256  & 256  & 128 & 256 & 128 & 256 & 128 & 128 & 64\\
         training epochs & 30  & 24  & 15  & 10 & 30 & 24 & 24 & 24 & 24 & 15\\
         warmup epochs & 5 & 5 & 2.5  & 2 & 5 & 5 & 5 & 5 & 5 & 2.5\\
         layer-wise lr decay~\cite{bao2021beit} & 0.7  & 0.8  & 0.7  & 0.8 & 0.7 & 0.85 & 0.7 & 0.85 & 0.7 & 0.85\\
         randaugment~\cite{randaugment} & \multicolumn{2}{c|}{(9, 0.5)} & \multicolumn{2}{c|}{(9, 0.5)} & \multicolumn{2}{c|}{(9, 0.5)} & \multicolumn{2}{c|}{(9, 0.5)} & \multicolumn{2}{c}{(9, 0.5)}\\ 
         label smoothing~\cite{szegedy2016inceptionv3} & \multicolumn{2}{c|}{0.1} & \multicolumn{2}{c|}{0.1} & \multicolumn{2}{c|}{0.1} & \multicolumn{2}{c|}{0.1} & \multicolumn{2}{c}{0.1}\\
         stochastic depth & \multicolumn{2}{c|}{-} & \multicolumn{2}{c|}{-} & \multicolumn{2}{c|}{-} & - & 0.2 & \multicolumn{2}{c}{-}\\
         \bottomrule
    \end{tabular}
    \caption{TAdaFormer training settings on K710, K400, SSV1/SSV2, and EK100.}
    \label{tab:training_tadaformer}
\end{table*}

\subsection{Action Localization}
We evaluate our model on the action localization task using two large-scale datasets. 
The overall pipeline for our action localization evaluation is divided into finetuning the classification models, obtaining action proposals, and classifying the proposals.

\textbf{Finetuning. }On \textit{Epic-Kitchens}, we simply use the evaluated action classification model. On \textit{HACS}, following~\cite{hacscompetition}, we initialize the model with Kinetics-400 pre-trained weights and train the model with adamW~\cite{adamw} for 30 epochs (8 warmups) using 32 GPUs. 
The mini-batch size is 16 videos per GPU. 
The base learning rate is set to 0.0002, with cosine learning rate decay as in Kinetics.
In our case, only the segments with action labels are used for training.

\textbf{Proposal generation.} 
For the action proposals, a boundary matching network (BMN)~\cite{bmn} is trained over the extracted features on the two datasets. 
On \textit{Epic-Kitchens}, we extract features with the videos uniformly decoded at 60 FPS. 
For each clip, we use 8 frames with an interval of 8 to be consistent with finetuning, which means a feature roughly covers a video clip of one seconds. 
The interval between each clip for feature extraction is 8 frames (\textit{i.e.,} 0.133 sec) as well.
The shorter side of the video is resized to 224 and we feed the whole spatial region into the backbone to retain as much information as possible.
Following \cite{ek100actionlocalization}, we generate proposals using BMN based on sliding windows.
The predictions on the overlapped region of different sliding windows are simply averaged.
On \textit{HACS}, the videos are decoded at 30 FPS, and extend the interval between clips to be 16 (\textit{i.e.,} 0.533 sec) because the actions in HACS last much longer than in Epic-Kitchens. 
The shorter side is resized to 128 for efficient processing. 
For the settings in generating proposals, we mainly follow~\cite{hacscompetition}, except that the temporal resolution is resized to 100 in our case instead of 200.

\textbf{Classification. }On \textit{Epic-Kitchens}, we classify the proposals with the fine-tuned model using 6 clips. Spatially, to comply with the feature extraction process, we resize the shorter side to 224 and feed the whole spatial region to the model for classification.
On \textit{HACS}, considering the property of the dataset that only one action category can exist in a video, we obtain the video level classification results by classifying the video level features, following~\cite{hacscompetition}.

\textbf{Action localization with ActionFormer. }
We follow all the settings in~\cite{actionformer-ek100-2022-challenge-report,zhang2022actionformer} for action localization experiments with ActionFormer.

\textbf{Evaluation.} For evaluation, we follow the standard evaluation protocol used in the respective datasets, \textit{i.e.,} the average mean Average Precision (average mAP) at IoU threshold [0.5:0.05:0.95] for HACS~\cite{hacs} and [0.1:0.1:0.5] for Epic-Kitchens-100~\cite{ek100}. 

\section{Model structures}
\label{appendix:modelstructure}
\newcommand{\blockrtd}[3]{\multirow{4}{*}{\(\left[\begin{array}{c}\text{1\x1$^\text{2}$, #2}\\[-.1em] \textbf{\textcolor{themebrown}{\text{3\x3$^\text{2}$, #2}}}\\[-.1em] \text{1\x1$^\text{2}$, #1}\end{array}\right]\)\x#3}
}
\newcommand{\blockrtpod}[3]{\multirow{4}{*}{\(\left[\begin{array}{c}\text{1\x1$^\text{2}$, #2}\\[-.1em] \textbf{\textcolor{themebrown}{\text{1\x3$^\text{2}$, #2}}}\\[-.1em] 
\textbf{\textcolor{themegreen}{\text{3\x1$^\text{2}$,#2}}}\\[-.1em]\text{1\x1$^\text{2}$, #1}\end{array}\right]\)\x#3}
}
\newcommand{\blockresnet}[3]{\multirow{4}{*}{\(\left[\begin{array}{c}\text{1\x1$^\text{2}$, #2}\\[-.1em] \textbf{\textcolor{themebrown}{\text{1\x3$^\text{2}$, #2}}}\\[-.1em] \text{1\x1$^\text{2}$, #1}\end{array}\right]\)\x#3}
}
\begin{table*}[t]
\centering
\caption{Model structure of R3D, R(2+1)D and R2D that we used in our experiments.
\textbf{\textcolor{themebrown}{Brown}} and \textbf{\textcolor{themegreen}{green}} fonts indicate respectively the default convolution operation and optional operation that can be replaced by TAdaConv. (Better viewed in color.)
}
\tablestyle{3pt}{1.08}
\begin{tabular}{c|c|c|c|c}
\shline
\bf Stage & \bf R3D & \bf R(2+1)D & \bf R2D & \bf output sizes \\
\shline
Sampling & interval 8, 1$^\text{2}$ & interval 8, 1$^\text{2}$ & interval 8, 1$^\text{2}$ &  8\x224\x224   \\
\hline
\multirow{2}{*}{conv$_1$} & 3\x7$^\text{2}$, {64} & 1\x7$^\text{2}$, {64} & 1\x7$^\text{2}$, {64} & \multirow{2}{*}{8\x112\x112}     \\
~ & stride 1, 2$^\text{2}$ & stride 1, 2$^\text{2}$ & stride 1, 2$^\text{2}$  \\
\hline
\multirow{4}{*}{res$_2$} & \blockrtd{{256}}{{64}}{3} & \blockrtpod{{256}}{{64}}{3} & \blockresnet{{256}}{{64}}{3} & \multirow{4}{*}{8\x56\x56} \\
&  &  & \\
&  &  & \\
&  &  & \\
\hline
\multirow{4}{*}{res$_3$} & \blockrtd{{512}}{{128}}{4} & \blockrtpod{{512}}{{128}}{4} & \blockresnet{{512}}{{128}}{4} & \multirow{4}{*}{8\x28\x28} \\
&  &  & \\
&  &  & \\
&  &  & \\
\hline
\multirow{4}{*}{res$_4$} & \blockrtd{{1024}}{{256}}{6} & \blockrtpod{{1024}}{{256}}{6} & \blockresnet{{1024}}{{256}}{6} & \multirow{4}{*}{8\x14\x14} \\
&  &  & \\
&  &  & \\
&  &  & \\
\hline
\multirow{4}{*}{res$_5$} & \blockrtd{{2048}}{{512}}{3} & \blockrtpod{{2048}}{{512}}{3} & \blockresnet{{2048}}{{512}}{3} & \multirow{4}{*}{8\x7\x7} \\
&  &  & \\
&  &  & \\
&  &  & \\
\hline
\multicolumn{4}{c|}{global average pool, fc}  & 1\x1\x1 \\
\shline
\end{tabular}
\label{tab:arch}
\end{table*}
The detailed model structures for R2D, R(2+1)D and R3D is specified in Table~\ref{tab:arch}. 
We highlight the convolutions that are replaced by TAdaConv by default or optionally.
For all of our models, a small modification is made in that we remove the max pooling layer after the first convolution and set the spatial stride of the second stage to be 2, following~\cite{corrnet}. 
Temporal resolution is kept unchanged following recent works~\cite{slowfast,tea,stm}. 
Our \textit{R3D} is obtained by simply expanding the R2D baseline in the temporal dimension by a factor of three. 
We initialize with weights reduced by 3 times, which means the original weight is evenly distributed in adjacent time steps.
We construct the \textit{R(2+1)D} by adding a temporal convolution operation after the spatial convolution. 
The temporal convolution can also be optionally replaced by TAdaConv, as shown in both the manuscript and Table~\ref{tab:pluginevalepickitchen}.
For its initialization, the temporal convolution weights are randomly initialized, while the others are initialized with the pre-trained weights on ImageNet.
For SlowFast models, we keep all the model structures identical to the original work~\cite{slowfast}.

For TAdaConvNeXt, we keep most of the model architectures as in ConvNeXt~\cite{convnext}, except that we use a tubelet embedding similar to~\cite{arnab2021vivit}, with a size of 3\x4\x4 and stride of 2\x4\x4. Center initialization is used as in \cite{arnab2021vivit}. Based on this, we simply replace the depth-wise convolutions with TAdaConv to construct TAdaConvNeXt. 
For TAdaConvNeXtV2, we additionally substitute TAdaConv for TAdaConvV2 and introduce the temporal aggregation scheme.
\begin{figure*}[t]
\centering
\includegraphics[width=\textwidth]{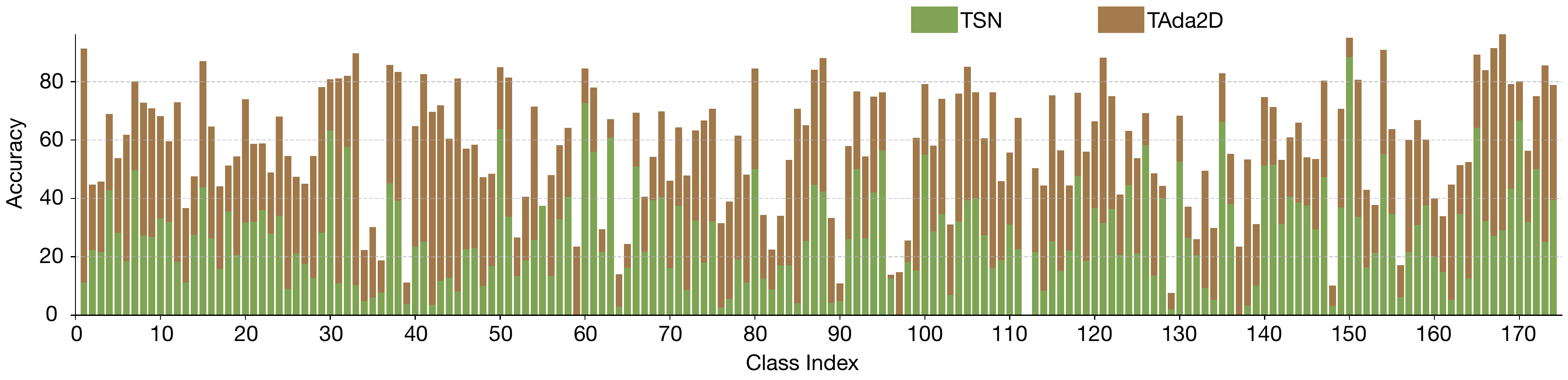}
\caption{\textbf{Per-category performance comparison of TAda2D against the baseline TSN.} We achieve an average per-category performance improvement of 30.35\%. }
\label{fig:diff_tsn}
\end{figure*}
\begin{figure*}[t]
\centering
\includegraphics[width=\textwidth]{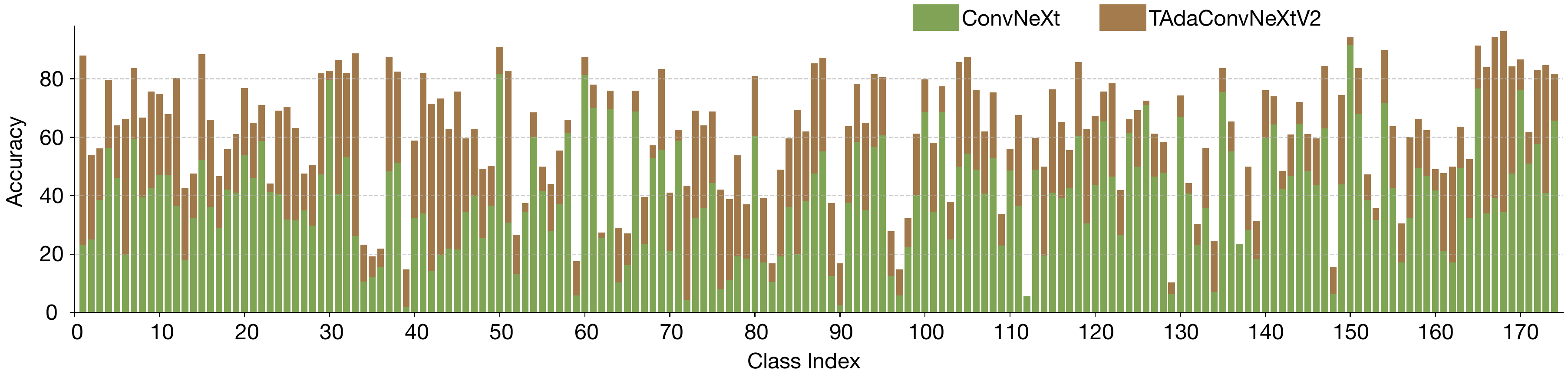}
\caption{\textbf{Per-category performance comparison of TAdaConvNeXtV2 against the baseline ConvNeXt.} We achieve an average per-category performance improvement of 21.36\%. }
\label{fig:diff_convnext}
\end{figure*}
\begin{table*}[t]
\caption{Ablation studies.}
\vspace{-3mm}
\centering
\subfloat[Ablation studies on kernel size with linear calibration weight generation function.
\label{tab:ablationstudieskernelsizelin}]{
\tablestyle{10pt}{1.0}
\begin{tabular}{cc}
\shline
\bf Kernel size & \bf Top-1\\
\hline
1 & 37.5 \\
3 & 56.5 \\
5 & 57.3 \\
7 & 56.5 \\
\shline
\end{tabular}
}\hspace{3mm}
\subfloat[Ablation studies on kernel size with non-linear calibration weight generation function.
\label{tab:ablationstudieskernelsizenonlin}]{
\tablestyle{4pt}{1.0}
\begin{tabular}{c|cccc}
\shline
~ & \bf K$_2$=1 & \bf K$_2$=3 & \bf K$_2$=5 & \bf K$_2$=7 \\
\hline
\bf K$_1$=1 & 36.8 & 57.1 & 57.8 & 57.9 \\
\bf K$_1$=3 & 57.3 & 57.8 & 57.9 & 58.0 \\
\bf K$_1$=5 & 57.6 & 57.9 & 58.2 & 57.9 \\
\bf K$_1$=7 & 57.4 & 57.6 & 58.0 & 57.6 \\
\shline
\end{tabular}
}\hspace{3mm}
\subfloat[Ablation studies on reduction ratio $r$ for $\mathbf{K_1}=\mathbf{K_2}=3$.
\label{tab:ablationstudiesreductionratio}]{
\tablestyle{6pt}{1.0}
\begin{tabular}{cc}
\shline
\bf Ratio $r$ & \bf Top-1 \\
\hline
1 & 57.79\\
2 & 57.83\\
4 & 57.78\\
8 & 57.66\\
\shline
\end{tabular}
}
\end{table*}
\begin{table*}[t]
\caption{Classification accuracy on Epic-Kitchens-100~\cite{ek100}. $\bm{\uparrow}$ indicates the main evaluation metric for the dataset. For fair comparison, we implement all the baseline models using our own training strategies.}
\centering
\tablestyle{3pt}{1.0}
\begin{tabular}{lccccccccc}
\toprule
~ & ~ & ~ & ~ & \multicolumn{3}{c}{\bf Top-1} & \multicolumn{3}{c}{\bf Top-5} \\
\cmidrule(r){5-7}\cmidrule(r){8-10}
\bf Model & \bf Frames & \bf GFLOPs & \bf Params.& \textbf{Act.}$\bm{\uparrow}$ & Verb & Noun &\textbf{Act.}$\bm{\uparrow}$ & Verb & Noun \\
\midrule
SlowFast 4\x16 & 4+32 & 36.10 & 34.5M & 38.17 & 63.54 & 48.79 & 58.68 & 89.75 & 73.37 \\
\graycell SlowFast 4\x16 + TAdaConv & \graycell4+32 & \graycell36.11 & \graycell37.7M & \graycell39.14 &\graycell 64.50 &\graycell 49.59 &\graycell 59.21 &\graycell 89.67 &\graycell 73.88 \\
\midrule
SlowFast 8\x8 & 8+32 & 65.71 & 34.5M & 40.08 & 65.05 & 50.72 & 60.10 & 90.04 & 74.26 \\ 
\graycell SlowFast 8\x8 + TAdaConv &\graycell 8+32 &\graycell 65.73 &\graycell 37.7M &\graycell 41.35 &\graycell 66.36 &\graycell 52.32 &\graycell 61.68 &\graycell 90.59 &\graycell 75.89 \\
\midrule
R(2+1)D & 8 & 49.55 & 28.1M & 37.45 & 62.92 & 48.27 & 58.02 & 89.75 & 73.60 \\
\graycell R(2+1)D + TAdaConv$_{\text{2d}}$ &\graycell 8 &\graycell 49.57 &\graycell 31.3M &\graycell 39.72 &\graycell 64.48 &\graycell 50.26 &\graycell 60.22 &\graycell 90.01 &\graycell 75.06\\
\graycell R(2+1)D + TAdaConv$_{\text{2d+1d}}$ &\graycell 8 &\graycell 49.58 &\graycell 34.4M &\graycell 40.10 &\graycell 64.77 &\graycell 50.28 &\graycell 60.45 &\graycell 89.99 &\graycell 75.55\\
\midrule
R3D & 8 & 84.23 & 47.0M & 36.67 & 61.92 & 47.87 & 57.47 & 89.02 & 73.05 \\
\graycell R3D + TAdaConv$_{\text{3d}}$ &\graycell 8 &\graycell 84.24 &\graycell 50.1M &\graycell 39.30 &\graycell 64.03 &\graycell 49.94 &\graycell 59.67 &\graycell 89.84 &\graycell 74.56\\
\bottomrule
\end{tabular}
\label{tab:pluginevalepickitchen}
\end{table*}
\newcommand{\hacs}[1]{\multirow{3}{*}{#1}}
\begin{table*}[t]
\caption{Plug-in evaluation of TAdaConv on the action localization on HACS and Epic-Kitchens. $\bm{\uparrow}$ indicates the main evaluation metric for the dataset. `S.F.' is SlowFast network.}
\centering
\tablestyle{2pt}{1.0}
\begin{tabular}{lccccccccccccc}
\toprule
~ & \multicolumn{6}{c}{\bf HACS} & \multicolumn{7}{c}{\bf Epic-Kitchen-100} \\
\cmidrule(r){2-7} \cmidrule(r){8-14}
\bf Model & @0.5 & @0.6 & @0.7 & @0.8 & @0.9 & \bf Avg.$\bm{\uparrow}$ & Task & @0.1 & @0.2 & @0.3 & @0.4 & @0.5 & \bf Avg.$\bm{\uparrow}$ \\
\midrule
\multirow{3}{*}{S.F. 8\x8} & \hacs{50.0} & \hacs{44.1} & \hacs{37.7} & \hacs{29.6} & \hacs{18.4} & \hacs{33.7} & Verb & 19.93 & 18.92 & 17.90 & 16.08 & 13.24 & 17.21 \\
~ & ~ & ~ & ~ & ~ & ~ & ~ & Noun & 17.93 & 16.83 & 15.53 & 13.68 & 11.41 & 15.07 \\
~ & ~ & ~ & ~ & ~ & ~ & ~ & \bf Act.$\bm{\uparrow}$ & 14.00 & 13.19 & 12.37 & 11.18 & 9.52 & 12.04\\
\midrule
\multirow{3}{*}{S.F. 8\x8 + TAdaConv} & \hacs{51.7} & \hacs{45.7} & \hacs{39.3} & \hacs{31.0} & \hacs{19.5} & \hacs{35.1} & Verb & 19.96 & 18.71 & 17.65 & 15.41 & 13.35 & 17.01 \\
~ & ~ & ~ & ~ & ~ & ~ & ~ & Noun & 20.17 & 18.90 & 17.58 & 15.83 & 13.18 & 17.13 \\
~ & ~ & ~ & ~ & ~ & ~ & ~ & \bf Act.$\bm{\uparrow}$ & 14.90 & 14.12 & 13.32 & 12.07 & 10.57 & 13.00\\
\bottomrule
\end{tabular}
\label{tab:pluginevallocalization}
\end{table*}

\section{Per-category improvement analysis on SSV2}
\label{appendix:percategoryimprovement}
This section provides a per-category improvement analysis on the Something-Something-V2 dataset in Fig.\ref{fig:diff_tsn} and Fig.~\ref{fig:diff_convnext}.
In terms of overall performance, our TAda2D achieves an improvement of 31.7\% over the baseline TSN, while TAdaConvNeXtV2 improves over ConvNeXt by 25.4\%. 
Our per-category analysis shows a mean improvement of 30.35\% and 21.36\% over all the classes, respectively for TAda2D and TAdaConvNeXtV2. 
Since both TSN and ConvNeXt have no temporal modeling capabilities, and our approach introduce similar modifications to the base model, the difference pattern in the per-category accuracy is similar. 
Hence, we take TAda2D as an example for analysis.
The largest improvement is observed in class 0 (78.5\%, \textit{Approaching something with your camera}), 32 (78.4\%, \textit{Moving away from something with your camera}), 30 (74.3\%, \textit{Lifting up one end of something without letting it drop down}), 44 (66.2\%, \textit{Moving something towards the camera}) and 41 (66.1\%, \textit{Moving something away from the camera}).
Most of these categories contain large movements across the whole video, whose improvement benefits from temporal reasoning over the global spatial context. 
For class 30, most of its actions last a long time (as it needs to be determined whether the end of something is let down or not). The improvements over the baseline mostly benefit from the global temporal context that is included in the weight generation process.

\section{Further ablation studies}
\label{appendix:ablation}
Here we provide further ablation studies on the kernel size in the calibration weight generation. As shown in Table~\ref{tab:ablationstudieskernelsizelin} and Table~\ref{tab:ablationstudieskernelsizenonlin}, kernel size does not affect the classification much, as long as the temporal context is considered. Further, Table~\ref{tab:ablationstudiesreductionratio} shows the sensitivity analysis on the reduction ratio, which demonstrate the robustness of our approach against different set of hyper-parameters.

\section{Further plug-in evaluation for TAdaConv on classification}
\label{appendix:pluginclassification}

In complement to the manuscript, we further show in Table~\ref{tab:pluginevalepickitchen} the plug-in evaluation on the action classification task on the Epic-Kitchens-100 dataset.
As in the plug-in evaluation on Kinetics and Something-Something-V2, we compare performances with and without TAdaConv over three baseline models, SlowFast~\cite{slowfast}, R(2+1)D~\cite{r21d} and R3D~\cite{retrace} respectively representing three kinds of temporal modeling techniques. 
The results are in line with our observation in the plug-in evaluation in the manuscript. Over all three kinds of temporal modelling strategies, adding TAdaConv further improves the recognition accuracy of the model.
\begin{figure*}[t]
\centering
\includegraphics[width=0.8\textwidth]{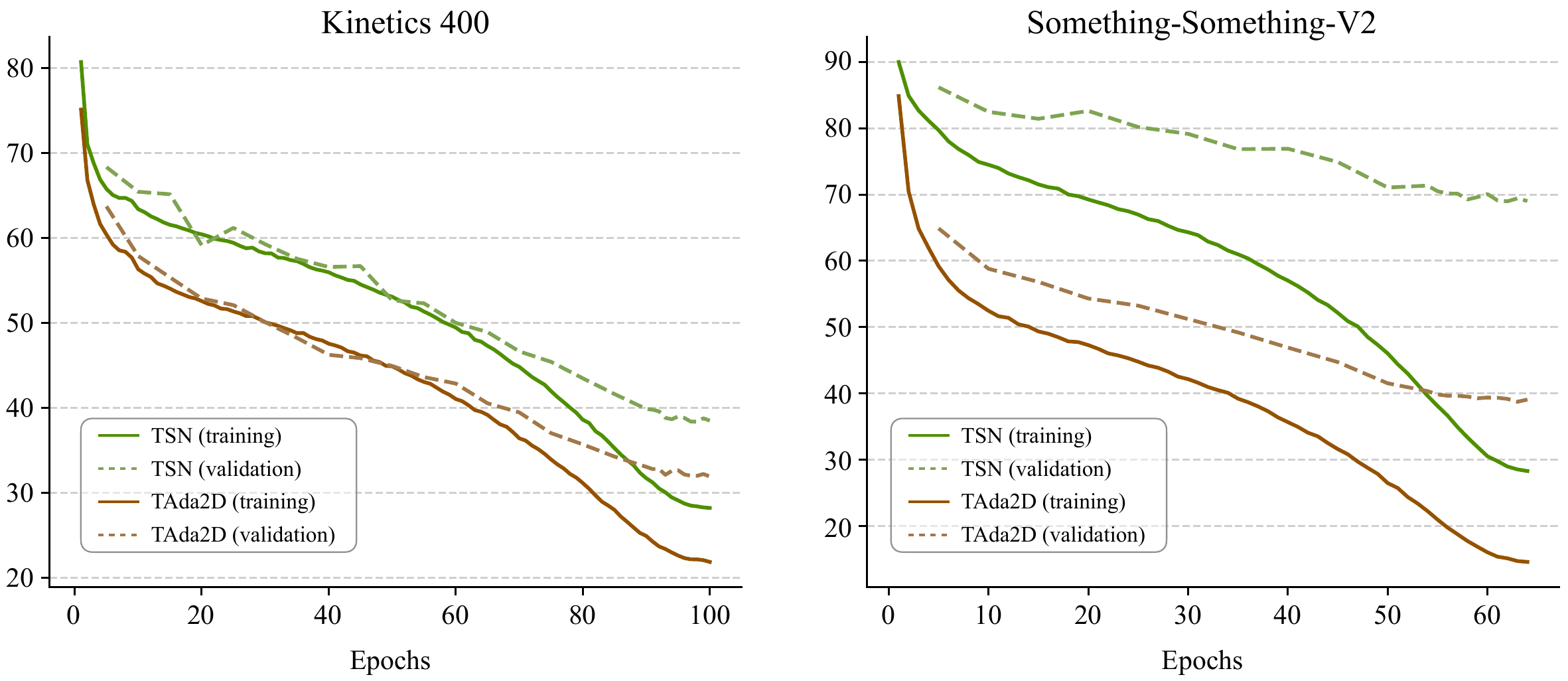}
\caption{\textbf{Training and validation on Kinetics-400 and Something-Something-V2 for TAda2D.} On both datasets, TAda2D shows a stronger capability of fitting the data and a better generality to the validation set. Further, TAda2D reduces the overfitting problem in Something-Something-V2.}
\label{fig:training_plot_tada2d}
\end{figure*}
\begin{figure*}[t]
\centering
\includegraphics[width=0.8\textwidth]{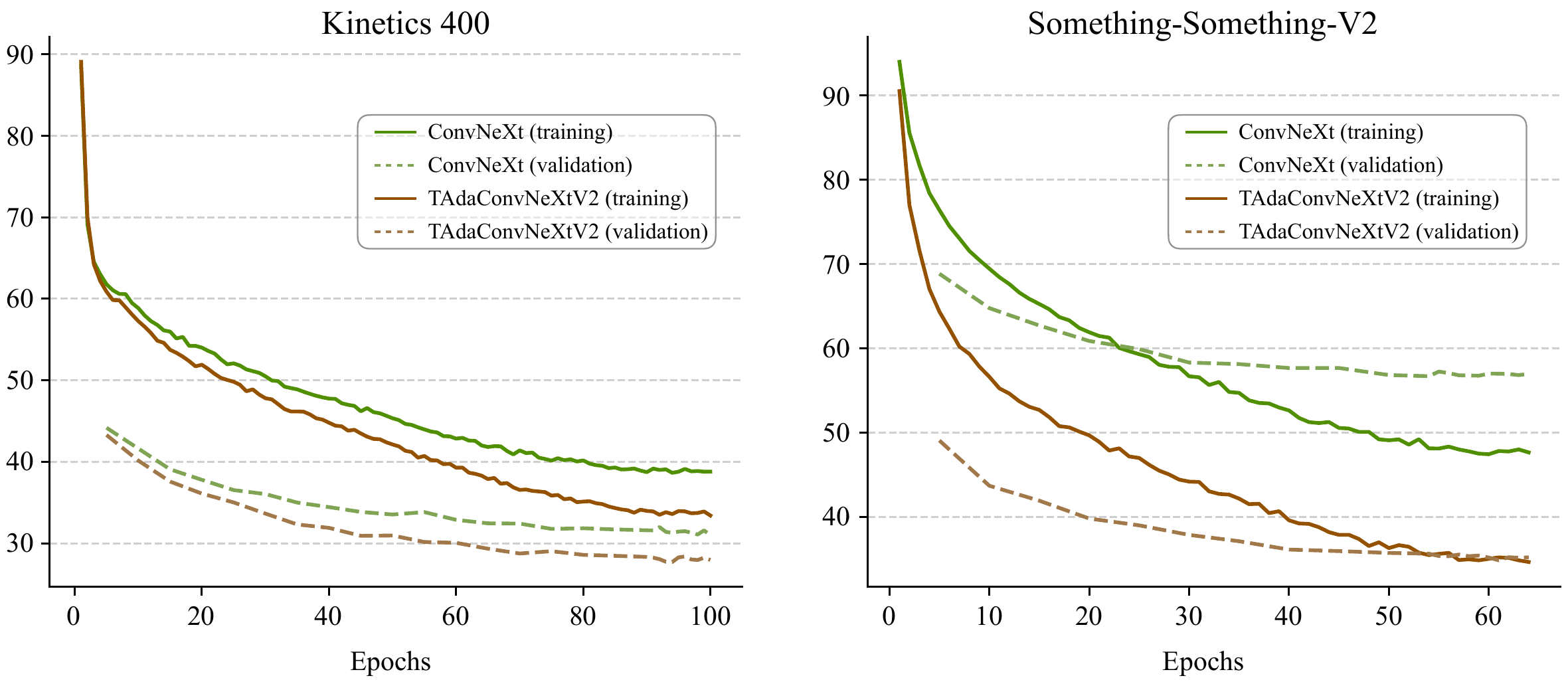}
\caption{\textbf{Training and validation on Kinetics-400 and Something-Something-V2 for TAdaConvNeXt.}}
\label{fig:training_plot_tadaconvnext}
\end{figure*}
\section{Plug-in evaluation for TAdaConv on Action Localization}
\label{appendix:pluginlocalization}
Here, we show the plug-in evaluation on the temporal action localization task. 
Specifically, we use SlowFast as our baseline, as it is shown to be superior in the localization performance in \cite{tcanet} compared to many early backbones.
The result is presented in Table~\ref{tab:pluginevallocalization}. 
With TAdaConv, the average mAP on HACS is improved by 1.4\%, and the average mAP on Epic-Kitchens-100 action localization is improved by 1.0\%. 

\section{Comparison of training procedure}
\label{appendix:trainingprocedure}
we compare the training procedure of TSN and TAda2D on Kinetics-400 and Something-Something-V2 in Fig.~\ref{fig:training_plot_tada2d}, and that of ConvNeXt and TAdaConvNeXtV2 in Fig.~\ref{fig:training_plot_tadaconvnext}. Although TAda2D and TAdaConvNeXtV2 are initialized to be identical to TSN and ConvNeXt, both TAda2D and TAdaConvNeXtV2 demonstrates a stronger performance on both training and validation sets. 

\begin{table*}[t]
    \tablestyle{3pt}{1.0}
\caption{Approach comparison between different dynamic filters. The weights column denotes how weights in respective approaches are obtained. The pre-trained weights colmun shows whether the weight generation can exploit pre-trained models such as ResNet~\cite{resnet}. 
}
\centering
\begin{tabular}{llccc}
\toprule
 ~ & ~ & \textbf{Temporal} &  \textbf{Location} & \textbf{Pretrained}\\
\textbf{Operations} & \textbf{Weights} & \textbf{Modelling} & \textbf{Adaptive} & \textbf{weights}\\
\midrule
 CondConv & Mixture of experts $\mathbf{W}=\sum_n f(\mathbf{x})_n\mathbf{W}_n$ &\xmark & \xmark & \xmark \\
 DynamicFilter & Completely generated $\mathbf{W}=g(\mathbf{x})$ &\xmark & \xmark & \xmark \\
 DDF & Completely generated $\mathbf{W}=g(\mathbf{x})$ & \xmark & \cmark & \xmark \\
 TAM & Completely generated $\mathbf{W}=g(\mathbf{x})$ & \cmark & \xmark & \xmark \\
 TAdaConv & Calibrated from a base weight $\mathbf{W}=h(\mathbf{x})\mathbf{W}_b$& \cmark & \cmark & \cmark \\
\bottomrule
\end{tabular}
\label{tab:compdyconvfull}
\end{table*}
\begin{table*}[t]
    \tablestyle{5pt}{1.0}
\caption{{Performance comparison with other dynamic filters.
\textit{Our Init.} denotes initializing the calibration weights to ones so that the initial calibrated weights is identical to the pre-trained weights. Temp. Varying is short for temporally varying, which indicates different weights for different temporal locations (frames).
* denotes that the branch was originally not designed for generating filter or calibration weights, but we slightly modified the structure so that it can be used for calibration weight generation. \textbf{\color{forestgreen}(Numbers in brackets)} show the performance improvement brought by our initialization scheme for calibration weights.}
}
\centering
\begin{tabular}{lcccl}
\toprule
\textbf{Calibration Generation} & \textbf{Our Init.} & \textbf{Temp. Varying} & \textbf{Generation source} & \textbf{Top-1} \\
\midrule
DynamicFilter & \xmark & \xmark & $\text{GAP}_{st}(\mathbf{x}) (C\times 1)$ & 41.7 \\
DDF-like & \xmark & \cmark & $\text{GAP}_{st}(\mathbf{x}) (C\times 1)$ & 49.8 \\
TAM (global branch) & \xmark & \xmark & $\text{GAP}_{s}(\mathbf{x}) (C\times T)$ & 39.7 \\
TAM (local*+global branch) & \xmark & \cmark & $\text{GAP}_{s}(\mathbf{x}) (C\times T)$ & 41.3 \\
\midrule
DynamicFilter & \cmark  & \xmark & $\text{GAP}_{st}(\mathbf{x}) (C\times 1)$ & 51.2 \textbf{\color{forestgreen}(+9.5)} \\
DDF-like & \cmark  & \cmark & $\text{GAP}_{st}(\mathbf{x}) (C\times 1)$ & 53.8 \textbf{\color{forestgreen}(+4.0)} \\
TAM (global branch) & \cmark & \xmark & $\text{GAP}_{s}(\mathbf{x}) (C\times T)$ & 52.9 \textbf{\color{forestgreen}(+13.2)}\\
TAM (local*+global branch) & \cmark & \cmark & $\text{GAP}_{s}(\mathbf{x}) (C\times T)$ & 54.3 \textbf{\color{forestgreen}(+13.0)}\\
\midrule
TAdaConv w/o global info $\mathbf{g}$ & \cmark & \cmark & $\text{GAP}_{s}(\mathbf{x}) (C\times T)$  & 57.9 \\
\midrule
\multirow{2}{*}{TAdaConv} & \multirow{2}{*}{\cmark} & \multirow{2}{*}{\cmark} & both $\text{GAP}_{st}(\mathbf{x}) (C\times 1)$  & \multirow{2}{*}{59.2} \\
~ & ~ & ~ & and $\text{GAP}_{s}(\mathbf{x}) (C\times T)$ & ~\\
\bottomrule
\end{tabular}
\label{tab:compdyconvperf}
\end{table*}

\section{Comparison with existing dynamic filters}
\label{appendix:comparisondynamicfilter}

In this section, we compare our TAdaConv with previous dynamic filters in two perspectives, respectively the difference in the methodology and in the performance.
\subsection{Comparison in terms of methodology}
We compare TAdaConv with several representative dynamic filtering approaches in image and in videos, respectively CondConv~\cite{condconv}, DynamicFilter~\cite{dynamicfilter}, DDF~\cite{ddf} and TAM~\cite{tam}.

The first difference in terms of methodology lies in the source of weights, where previous approaches obtain weights by mixture of experts or generation completely dependent on the input.
\textit{Mixture of experts} denotes $\mathbf{W}=\sum_n\alpha_n\mathbf{W}_n$, where $\alpha_n$ is a scalar obtained by a function $f$, \textit{i.e.,} $\mathbf{W}=\sum_n f(\mathbf{x})_n\mathbf{W}_n$.
\textit{Completely generated} means the weights are only dependent on the input, \textit{i.e.,} $\mathbf{W}=g(\mathbf(\mathbf{x}))$, where $g$ generates complete kernel for the convolution.
In comparison, the weights in TAdaConv are obtained by \textit{calibration}, \textit{i.e,,} $\mathbf{W}=\bm{\alpha}\mathbf{W}_b$, where $\bm{\alpha}$ is a vector calibration weight and $\bm{\alpha}=h(\mathbf(\mathbf{x}))$ where $h(.)$ generates the calibration vector for the convolutions.
Hence, this fundamental difference in how to obtain the convolution weights makes the previous approaches difficult to exploit pre-trained weights, while TAdaConv can easily load pre-trained weights in $\mathbf{W}_b$.
This ability is essential for video models to speed up the convergence.

The second difference lies in the ability to perform temporal modelling. The ability to perform temporal modelling does not only mean the ability to generate weights according to the whole sequence in dynamic filters for videos, but it also requires the model to generate different weights for the same set of frames with different orders.
For example, weights generated by the global descriptor obtained by global average pooling over the whole video $\text{GAP}_{st}$ does not have the temporal modelling ability, since they can not generate different weights if the order of the frames in the input sequence are reversed or randomized. Hence, most image based approaches based on global descriptor vectors (such as CondConv and DynamicFilter) or based on adjacent spatial contents (DDF) can not achieve temporal modelling. TAM generates convolution weights for temporal convolutions based on temporally local descriptors obtained by the global average pooling over the spatial dimension $\text{GAP}_{s}$, which yields different weights if the sequence changes. Hence, in this sense, TAM has the temporal modelling abilities. In contrast, TAdaConv exploits both temporally local and global descriptors to utilize not only local but also global temporal contexts. Details on the source of the weight generation process is also shown in Table~\ref{tab:compdyconvperf}.

The third difference lies in whether the weights generated are shared for different locations. 
For CondConv, DynamicFilter and TAM, their generated weights are shared for all locations, while for DDF, the weights are varied according to spatial locations. In comparison, TAdaConv generates temporally adaptive weights.

\subsection{Comparison in the performance level}
Since TAdaConv is fundamentally different from previous approaches in the generation of calibration weights, it is difficult to directly compare the performance on video modelling, especially for those that are not designed for video modelling. 
However, since the calibration weight in TAdaConv $\bm{\alpha}$ is completely generated, \textit{i.e.,} $\bm{\alpha}=f(\mathbf(\mathbf{x}))$, we can use other dynamic filters to generate the calibration weights for TAdaConv. 
Since MoE-based approaches such as CondConv were essentially designed for applications with less memory constraint but high computation requirements, it is not suitable for video applications since it would be too memory-heavy for video models. 
Hence, we apply approaches that generate complete kernel weights to generate calibration weights and compare them with TAdaConv. 
The performance is listed in Table~\ref{tab:compdyconvperf}.

It is worth noting that these approaches originally generate weights that are randomly initialized. 
However, as is shown in the manuscript, our initialization strategy for the calibration weights are essential for yielding reasonable results, we further apply our initialization on these existing approaches to see whether their generation function is better than the one in TAdaConv.
In the following paragraphs, we provide details for applying representative previous dynamic filters in TAdaConv to generate the calibration weight.

For DynamicFilter~\cite{dynamicfilter}, the calibration weight $\bm{\alpha}$ is generated using an MLP over the global descriptor that is obtained by performing global average pooling over the whole input $\text{GAP}_{st}$, \textit{i.e.,} $\bm{\alpha}=\text{MLP}(\text{GAP}_{st}(\mathbf{x}))$.
In this case, the calibration weights are shared between different time steps. 

For DDF~\cite{ddf}, we only use the channel branch since it is shown in the manuscript that it is better to leave the spatial structure unchanged for the base kernel.
Similarly, the weights in DDF are also generated by applying an MLP over the global descriptor, \textit{i.e.,} $\bm{\alpha}=\text{MLP}(\text{GAP}_{st}(\mathbf{x}))$.
The difference between DDF and DynamicFilter is that for different time step, DDF generates a different calibration weight.

The original structure of TAM~\cite{tam} only generates kernel weights with its global branch and uses the local branch to generate attention maps over different time steps. 
In our experiments, we modify the TAM a little bit and further make the local branch generate kernel calibration weights as well. 
Hence, for the only-global version of TAM, the calibration weights are calculated as follows: $\bm{\alpha}=\mathcal{G}(\text{GAP}_{s}(\mathbf{x}))$, where $\text{GAP}_{s}$ denotes global average pooling over the spatial dimension and $\mathcal{G}$ denotes the global branch in TAM.
In this case, calibration weights are shared for all temporal locations.
For local+global version of TAM, the calibration weight are calculated by combining the results of the local $\mathcal{L}$ and the global branch $\mathcal{G}$, \textit{i.e.,} $\bm{\alpha}=\mathcal{G}(\text{GAP}_{s}(\mathbf{x}))\cdot\mathcal{L}(\text{GAP}_{s}(\mathbf{x}))$, where $\cdot$ denotes element-wise multiplication with broadcasting.
This means in this case, the calibration weights are temporally adaptive. Note that this is our modified version of TAM. The original TAM droes not have temporally adaptive convolution weights.

The results in Table~\ref{tab:compdyconvperf} show that (a) without our initialization strategy, previous approaches that generate random weights at initialization are not suitable for generating the calibration weights in TAdaConv; (b) our initialization strategy can conveniently change this and make previous approaches yield reasonable performance when they are used for generating calibration weights; and (c) the calibration weight generation function in TAdaConv, which combines the local and global context, outperform all previous approaches for calibration.

Further, when we compare TAdaConv without global information with TAM (local*+global branch), it can be seen that although both approach generates temporally varying weights from the frame descriptors $\text{GAP}_{s}(\mathbf{x})$ with shape $C\times T$, our TAdaConv achieves a notably higer performance. Adding the global information enables TAdaConv to achieve a more notable lead in the comparison with previous dynamic filters.


\end{document}